\documentclass{article}
\usepackage{arxiv}

\title{Synthetic Mixed Training:\\Scaling Parametric Knowledge Acquisition Beyond RAG}

\author[1]{Seungju Han}
\author[1]{Konwoo Kim}
\author[2]{Chanwoo Park}
\author[3]{Benjamin Newman}
\author[1]{Suhas Kotha}
\author[3]{Jaehun Jung}
\author[1]{James Zou}
\author[1]{Yejin Choi}

\affil[1]{Stanford University}
\affil[2]{MIT}
\affil[3]{University of Washington}

\begin{document}
\maketitle
\newcommand{\konwoo}[1]{\textcolor[HTML]{03A1FC}{[Konwoo: #1]}}
\newcommand{\sj}[1]{\textcolor{purple}{[SJ: #1]}}
\newcommand{\jaehun}[1]{\textcolor{blue}{[Jaehun: #1]}}
\newcommand{\suhas}[1]{\textcolor[HTML]{7030A0}{[Suhas: #1]}}
\newcommand{\yejin}[1]{\textcolor{red}{[yejin: #1]}}
\newcommand{\ben}[1]{\textcolor[HTML]{134f5c}{[ben: #1]}}

\begin{abstract}
Synthetic data augmentation helps language models learn new knowledge in data-constrained domains. However, naively scaling existing synthetic data methods by training on more synthetic tokens or using stronger generators yields diminishing returns below the performance of RAG. To break the RAG ceiling, we introduce Synthetic Mixed Training, which combines synthetic QAs and synthetic documents. This leverages their complementary training signals, and enables log-linear improvements as both synthetic data volume and generator strength increase. This allows the model to outperform RAG by a 2.6\% relative gain on QuaLITY, a long-document reading comprehension benchmark. In addition, we introduce Focal Rewriting, a simple technique for synthetic document generation that explicitly conditions document generation on specific questions, improving the diversity of synthetic documents and yielding a steeper log-linear scaling curve. On QuaLITY, our final recipe trains a Llama 8B model that outperforms RAG by 4.4\% relatively. Across models and benchmarks (QuaLITY, LongHealth, FinanceBench), our training enables models to beat RAG in five of six settings, outperforms by 2.6\%, and achieves a 9.1\% gain when combined with RAG.
\end{abstract}

\section{Introduction}

Language models fail to internalize all knowledge during pretraining, so recent studies have investigated whether domain-specific fine-tuning can improve knowledge learning. They report that retrieval-augmented generation (RAG)---the de facto approach for data-constrained domains---sets a strong upper bound that is difficult to surpass~\citep{ovadia2024fine, soudani2024fine}. This is because incorporating new knowledge into language model parameters is challenging in data-constrained settings. One common approach is to perform continued pretraining using synthetic data generated from domain-specific source documents. However, the vast majority of studies have found only limited success~\citep{yangsynthetic, lin2025learning, cacciatraining, eyuboglu2025cartridges, lampinen2025latent}. While it is natural to attribute this failure to the quality of the synthetic data generators, prior work has observed that stronger generators yield diminishing returns~\citep{lin2025learning, maini2025beyondweb, kang2025demystifying, maini2024rephrasing, niklaus2026_the_synthetic_data_playbook_generating_trillions_of_the_finest_tokens}.
In this work, we address these issues by answering the question:

\textit{\\How can we design synthetic data recipes for knowledge learning that scales better with number of synthetic tokens data and stronger generator?\\} 

\begin{figure}[t!]
  \centering
  \includegraphics[width=0.85\linewidth]{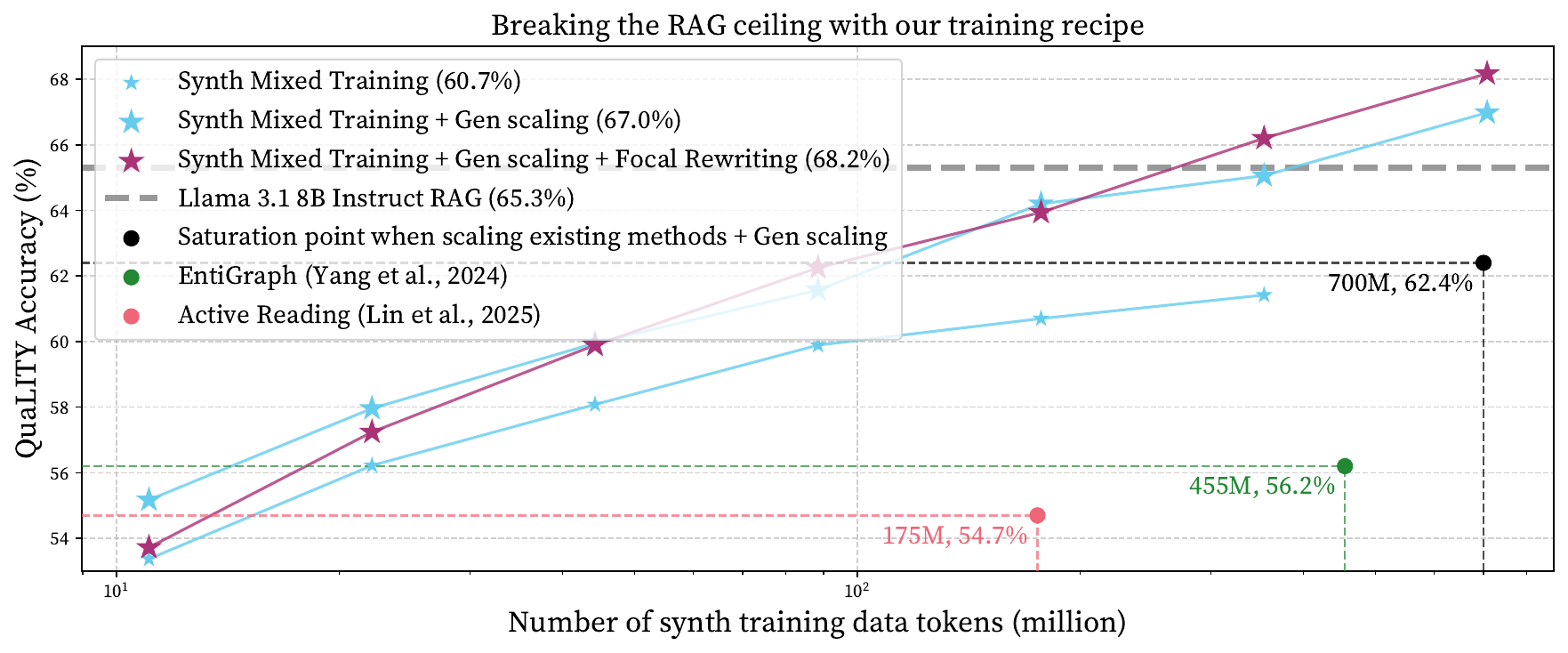}
  \label{fig:our-highlight}
  \vspace{-1em}
  \caption{\textbf{Naively scaling synthetic data plateaus, but our simple methods allows effective scaling and surpass RAG.} We evaluate four synthetic data generation strategies using both 8B and 70B generators, scaling training data up to 700M tokens. Across all four baselines, performance saturates and remains below RAG, showing that simply increasing synthetic data or compute is insufficient. In contrast, our two simple techniques---\textbf{Synthetic Mixed Training} and \textbf{Focal Rewriting}---exhibit clear log-linear scaling with both more data and a stronger generator, ultimately surpassing RAG. }
  \vspace{-1em}
  \label{fig:highlight}
\end{figure}

First, we investigate existing data generation algorithms and find that they do not scale well. We experiment with four existing data generation algorithms---one for generating synthetic QA pairs and three for generating synthetic documents---and use Llama 3.1 8B and 70B models to generate up to 700M synthetic tokens for training an 8B model. We find that, on a reading comprehension benchmark requiring the acquisition of new knowledge (QuaLITY;~\citet{pang-etal-2022-quality}), \textbf{existing training recipes based on synthetic data are insufficient to train a model that outperforms RAG, even when the data is generated by a 70B model that is much stronger than the 8B model being trained}. In particular, we find that training on synthetic QAs performs better than training on synthetic documents for a fixed generator. However, the gains from scaling the generator depend on the choice of data generation algorithm---document generation benefits more from generator scaling. Still, all methods exhibit diminishing returns as the number of synthetic tokens increases and remain 4.6\% behind RAG in relative accuracy.

Building on the observations that synthetic QAs and documents have different scaling properties with respect to data and generator strength, we hypothesize that QA and document data each provide unique benefits during training. To achieve the best of both worlds, we propose \textbf{Synthetic Mixed Training}, which combines synthetic QAs with synthetic documents during training. This substantially improves synthetic token efficiency, exhibits clear log-linear scaling behavior up to 700M synthetic tokens, and enables the model to surpass RAG by a 2.6\% relative gain. In addition, we introduce \textbf{Focal Rewriting}, a simple technique for diversifying the topics covered by synthetic documents by explicitly conditioning document generation on synthetic questions about the source document. This increases the lexical and semantic diversity of the synthetic documents and further improves model performance, yielding a 4.4\% relative accuracy gain over RAG.

We show that our recipe generalizes well across different base models, including Qwen3 models ranging from 1.7B to 14B parameters, and across three benchmarks: QuaLITY, LongHealth, and FinanceBench. In particular, our recipe enables models to outperform RAG in five of six setups, yielding an average relative accuracy gain of 2.6\% over vanilla RAG. Our approach is also complementary to RAG, providing a 9.1\% relative accuracy improvement over vanilla RAG. These results suggest an untapped potential of synthetic data in enhancing internalization of new knowledge for language models.

\section{Existing synthetic data recipes plateau when scaled}\label{sec:prelim}

Because the community lacks a holistic comparison of diverse synthetic data augmentation strategies, we present one here at scale in a controlled setup. Specifically, we present experimental results for training an 8B model on variants of synthetic data derived from documents in the QuaLITY benchmark~\citep{pang-etal-2022-quality}. To study how performance changes with data scale, we vary the number of synthetic tokens across runs, scaling up to 700M tokens.
\begin{figure}[t!]
  \centering
  \begin{subfigure}[t]{0.49\columnwidth}
    \centering
    \includegraphics[width=\linewidth]{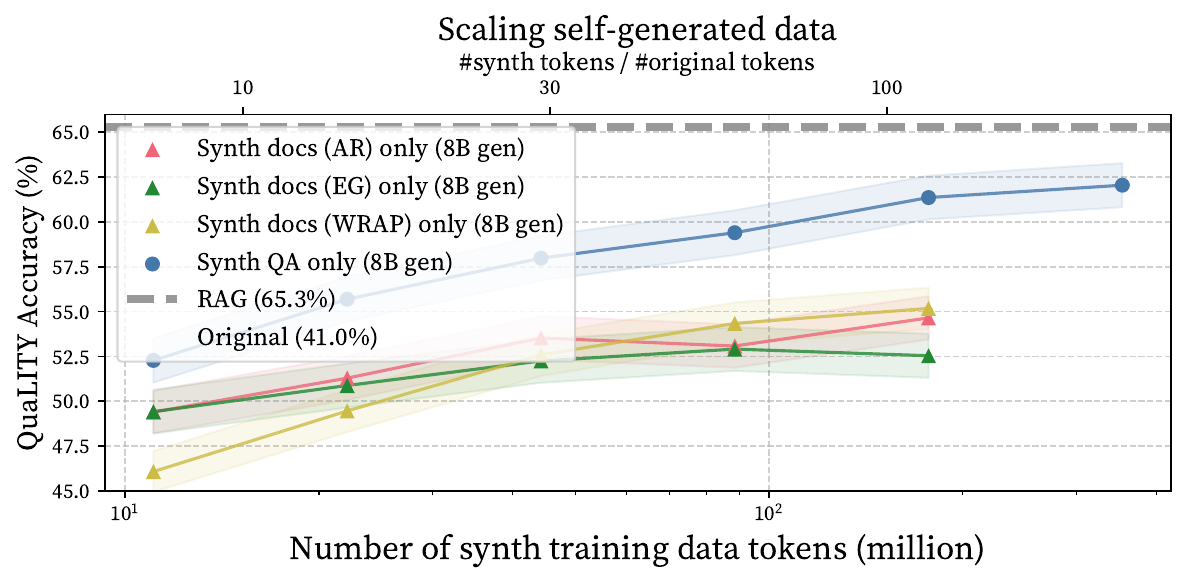}
    \label{fig:qa-vs-doc}
  \end{subfigure}\hfill
  \begin{subfigure}[t]{0.49\columnwidth}
    \centering
    \includegraphics[width=\linewidth]{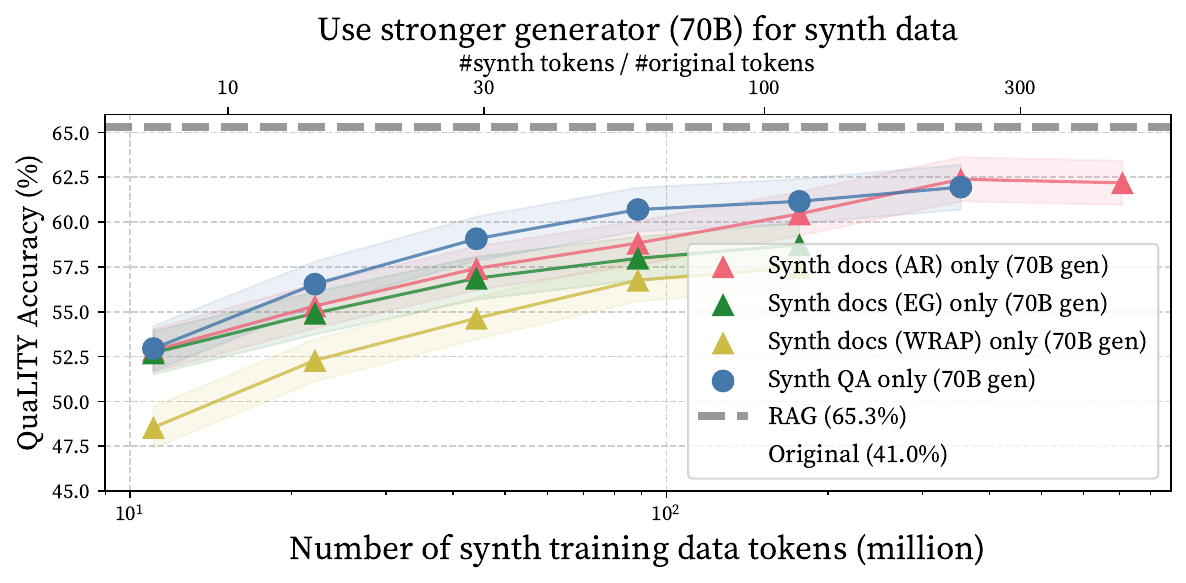}
    \label{fig:scaling-generator}
  \end{subfigure}
    \caption{(Left) \textbf{Comparing the data scaling of existing methods: self-generated (8B) synthetic QAs and synthetic documents.} This shows QuaLITY accuracy as a function of the number of synthetic training tokens; shaded areas indicate the standard deviation corresponding to the 95\% confidence interval, estimated from $n=8$ inference runs. We use Llama 3.1 8B Inst for both data generation and model training. AR indicates Active Reading~\citep{lin2025learning}, EG indicates EntiGraph~\citep{yangsynthetic}, and WRAP indicates rephrasing~\citep{maini2024rephrasing}. On QuALITY, synth QA is substantially more efficient than all existing methods which generate synthetic documents.
    (Right) \textbf{Scaling the generator to improve synthetic token efficiency.} 
    (1) Scaling the generator to 70B does not improve synth token efficiency for QA, only 0.1\% gain over 8B generator at 88M tokens. (2) In contrast, document-based methods do benefit from scaling the generator, achieving 4.5\% gain on average. (3) For all methods, even with a stronger generator, data scaling plateaus.
    }
  \label{fig:prelim}
\end{figure}

\paragraph{Setup.} 
We train the Llama 3.1 8B Instruct model~\citep{grattafiori2024llama} on the synthetic data using a fixed set of hyperparameters (except for learning rate; we search across three LRs---5e-6, 1e-6 and 5e-5---and report the best accuracy), varying only the method used to generate the synthetic data. Similar to \citet{yangsynthetic}, we use FineWeb~\citep{penedo2024fineweb} as a replay data, with a mixing ratio of 10\%. For evaluation, we use the QuaLITY multiple-choice QA set with a zero-shot instruction that asks the LM to provide a short explanation followed by the final answer.  See Appendix~\ref{sec:appendix-training} and \ref{sec:appendix-eval} for more details.

\paragraph{Synthetic document generation.} We test four data generation methods. All of these methods take original source documents as input and generate data grounded in them. For synthetic document generation, we test three algorithms. The first is rephrasing (WRAP;~\citet{maini2024rephrasing}), which rephrases documents using an LM. The second is EntiGraph (EG;~\citet{yangsynthetic}), which uses a two-stage approach: it first extracts core entities from the document using an LM, and then constructs documents describing the relationships between those entities using an LM again. The last is Active Reading (AR;~\citet{lin2025learning}), which also uses a two-stage approach--first, an LM generates strategies to rephrase the document, which are then incorporated into the instruction to guide the LM in rewriting the document. 

\paragraph{Synthetic QA generation.} For synthetic QA generation, we first generate diverse QA pairs using an LM with a simple instruction, and then use an LM again to produce a response with a short explanation for each question. This is similar to the QA generation approaches of \citet{lin2025learning} and \citet{yangsynthetic}, except that we explicitly instruct the LMs to generate explanations in the responses. We generate open-ended QA pairs rather than multiple-choice QA (MCQA) pairs, since generating difficult yet faithful distractors for MCQA is nontrivial. See Appendix~\ref{sec:appendix-gen} for more details.

\subsection{Scaling self-generated synthetic data has diminishing returns}

Figure~\ref{fig:prelim} (left) shows the evaluation results on QuaLITY when we train the model on synthetic data generated by the same 8B model (i.e., self-generated data). Training on synthetic QAs is more synthetic token efficient than training on any of the three synthetic document variants, while scaling synthetic documents begins to plateau. This was unexpected, as 
prior work found a different empirical result when training an 8B model with 8B-generated data~\citep{lin2025learning}. 
We speculate that the difference arises from their 
synthetic QA pairs containing only short answers without explanations, as we find that training the model on responses containing only the answer, without an explanation (i.e., using the answers from the initially generated QA pairs), leads to poor accuracy. However, when we specifically scale synthetic QAs up to 350M tokens, it also begins to show signs of plateauing. 

\subsection{Strategy matters when scaling generator to improve synthetic token efficiency}\label{subsec:prelim2}

We further explore a natural direction for improving synthetic token efficiency: scaling the generator model (use Llama 3.1 70B Instruct) to produce higher-quality synthetic data. Figure~\ref{fig:prelim} (right) shows the results of generator scaling for synthetic data generation. For synthetic QA generation, a stronger generator does not yield meaningful improvements in synthetic token efficiency, and performance also saturates as we scale the number of synthetic QA tokens. In contrast, for document generation, a stronger generator improves synthetic token efficiency: it closes the gap between training on synthetic QAs and training on synthetic documents. The choice of data generation algorithm also matters greatly: AR shows the best synthetic token efficiency among the variants when using the 70B generator, whereas WRAP shows the worst synthetic token efficiency relative to its performance with the 8B generator. However, 70B-generated AR documents also shows saturating performance when scaled up to 700M tokens.

The results suggest that \textbf{we should think carefully about which generation procedures are likely to improve with more capable generators}, as not every data generation algorithm benefits from a stronger generator model. Prior synthetic data works report related empirical findings that stronger generators do not always provide better synthetic data (without clear explanation): \citet{lin2025learning} shows that using 70B-generated data to train the 8B model underperforms compared to
using 8B generated data;
\citet{guha2025openthoughts} shows that a weaker, smaller generator (QwQ-32B) can produce better synthetic data for math and code than a stronger, larger generator (DeepSeek-R1);
and \citet{maini2025beyondweb} observes saturating returns when scaling the generator from a 3B to an 8B model for document rephrasing in pretraining setups. \citet{kang2025demystifying} shows that using a 70B model does not yield better performance when rephrasing pre-training data, and more recently, \citet{niklaus2026_the_synthetic_data_playbook_generating_trillions_of_the_finest_tokens} shows that a 1B generator is sufficient for rephrasing, supported by extensive experimental results.

\begin{figure}[t]
\centering
\includegraphics[width=0.8\linewidth]{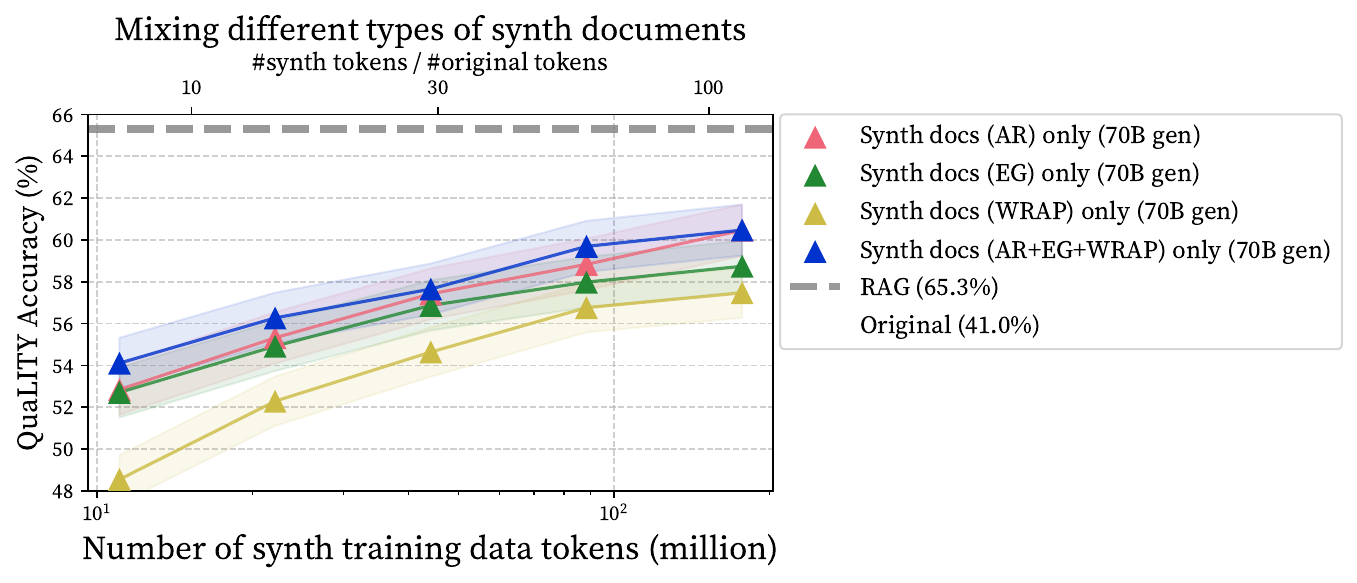}
\caption{\textbf{Mixing synthetic documents does not help.} Mixing different kinds of synthetic documents (blue line) provides a minimal gain over just using AR documents (pink line).}
\label{fig:document-mixing}
\vspace{-1em}
\end{figure}

\begin{figure}[t]
\centering
\includegraphics[width=0.8\linewidth]{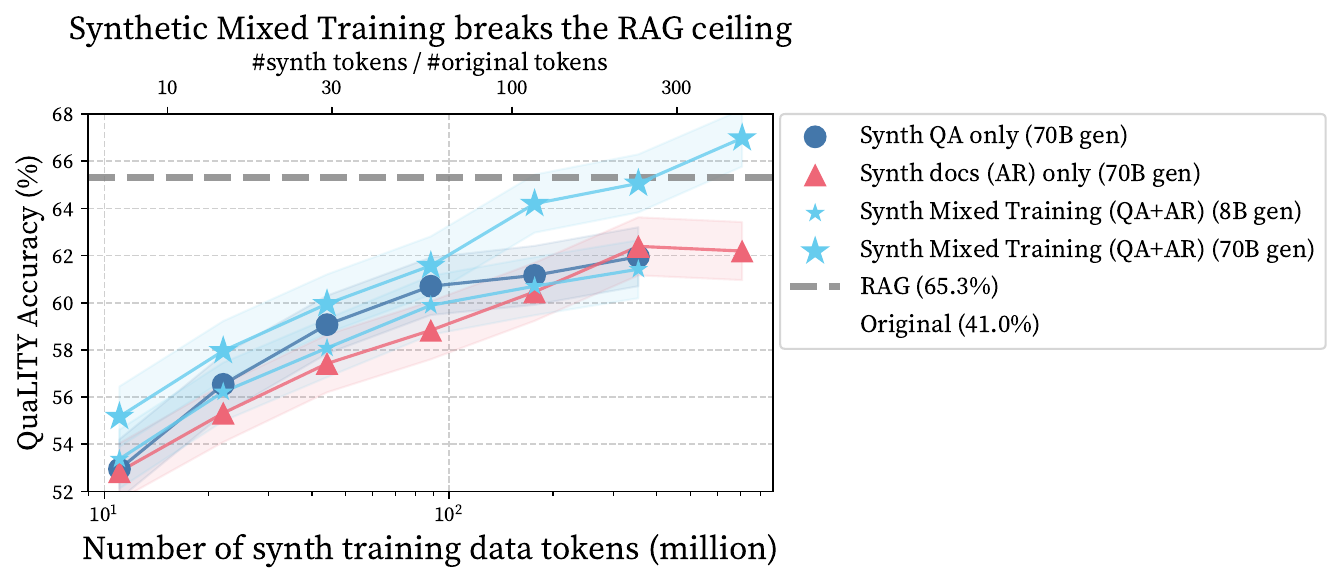}
\caption{\textbf{Synthetic Mixed Training breaks the RAG ceiling.} We combine 70B-generated synthetic QAs and AR documents at a 1:1 ratio, attempting to achieve the best of both worlds. This (skyblue line) yields performance comparable to RAG at 350M synthetic training tokens and ultimately surpasses RAG when scaled to 700M tokens.}
\label{fig:synth-mixed-training}

  \vspace{-1em}
\end{figure}

\section{Unlocking synthetic data scaling}

Building on our empirical findings, we present two simple methods for improving the efficiency of synthetic data and overcoming the limitations of existing synthetic data scaling. Our methods enable clear log-linear scaling up to 700M training tokens, and our trained 8B model substantially outperforms RAG on QuaLITY, achieving a 4.4\% relative accuracy gain. In addition, we find that these methods also work well with different base models (Qwen3 1.7B--14B) and additional benchmarks (LongHealth, FinanceBench).

\subsection{Synthetic Mixed Training: Mixing synthetic QAs and synthetic documents}

Since we have shown synthetic data strategies have different scaling properties, we investigate whether we can design a synthetic data recipe that achieves the best of all worlds.

We hypothesize that synthetic QAs and synthetic documents play different roles: synthetic QAs primarily teach \textit{behavioral knowledge} 
(e.g., how to recall facts through chain-of-thought reasoning), which can transfer across domains, whereas synthetic documents primarily teach \textit{factual knowledge}, which is more domain-specific. 
To validate this hypothesis, we test mixing all three types of synthetic documents during training, and Figure~\ref{fig:document-mixing} shows that the synthetic document types are similar and provide minimal improvement when mixed.
Additionally, we measure the similarity between data points in gradient space~\citep{jungprismatic} to quantify how synthetic QAs and documents from different domains are similar, see Appendix~\ref{sec:appendix-grad} for the analysis on this.

Based on this hypothesis, we explore training models with a mixture of synthetic QAs and synthetic documents. We call this approach \textbf{Synthetic Mixed Training}.\footnote{This name is inspired by Mixed Training, introduced by \citet{allen2024physics}. Their approach is designed for pretraining from scratch and does not consider synthetic documents when mixing.} We choose AR for document generation because it benefits the most from generator scaling, and we mix QAs and AR documents at a 1:1 ratio. Figure~\ref{fig:synth-mixed-training} shows the results of Synthetic Mixed Training using 70B-generated QA and AR documents, compared with training only on synthetic QAs or only on AR documents. The scaling curve exhibits persistent log-linear behavior as the number of training tokens increases up to 700M, eventually surpassing RAG with 67.0\% accuracy (+2.6\% relative gain). Moreover, when comparing results using 70B- and 8B-generated data for Synthetic Mixed Training, we find that a stronger generator is substantially more helpful. This suggests that QA simply dominates AR at 8B generated scale, highlighting AR’s unique benefit from generator scaling.

\begin{figure}[t!]
\centering
\includegraphics[width=0.9\linewidth]{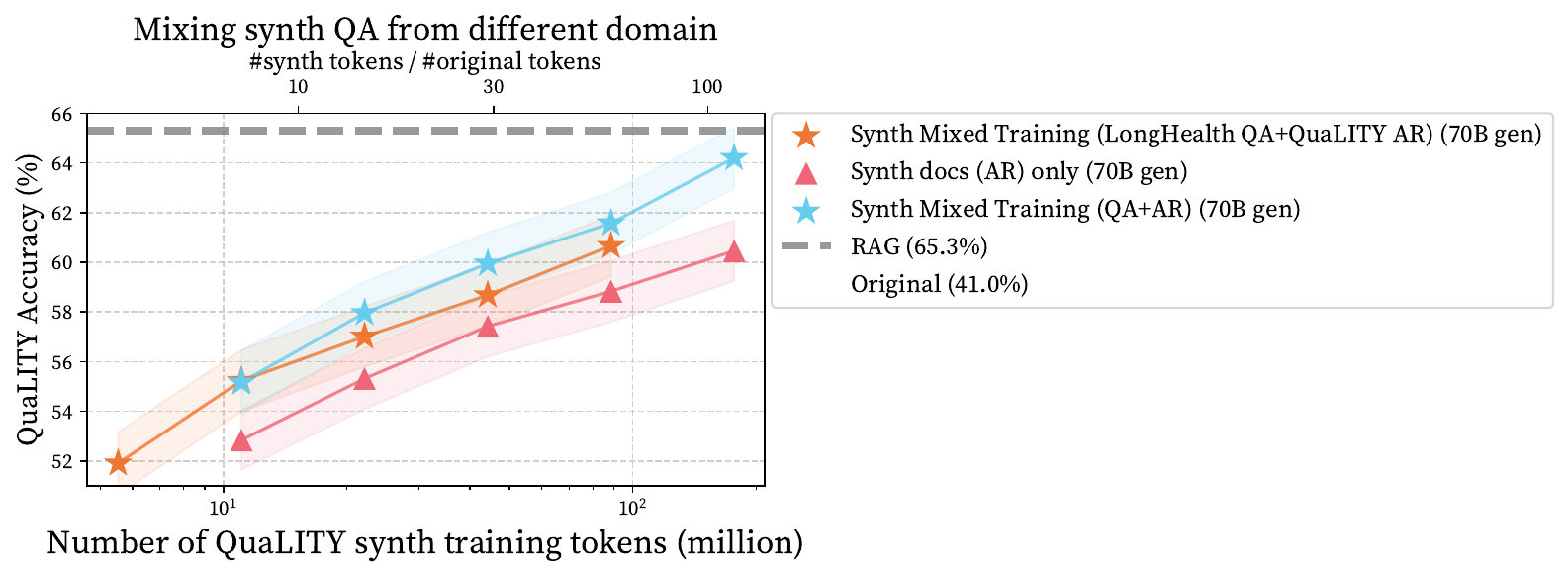}
\caption{\textbf{Synthetic Mixed Training with mixture of domains.} Here, the x-axis denotes the number of synthetic tokens grounded in the QuaLITY dataset. (1) Mixing 50\% synthetic QAs grounded in a different domain with 50\% synthetic documents grounded in the target domain yields a better scaling curve than training solely on target-domain synthetic documents. 
(2) The best performance comes from mixing target-domain synthetic QAs with target-domain synthetic documents, suggesting that synthetic QAs not only teaches recall behavior but also provides domain-specific knowledge.
}
\label{fig:cross-domain-qa-doc-mixing}
\end{figure}

We further test whether target-domain synthetic QAs are important by mixing synthetic QAs from a different dataset (LongHealth, an unrelated domain) with synthetic AR documents from QuaLITY (the target domain). Figure~\ref{fig:cross-domain-qa-doc-mixing} shows two findings. (1) Synthetic QAs can teach domain-agnostic behavior that is difficult to learn from AR documents alone: mixing LongHealth synthetic QAs with QuaLITY synthetic AR documents improves synthetic token efficiency on QuaLITY compared to using only QuaLITY synthetic AR documents. This supports our hypothesis and helps explaining the synergy between synthetic QAs and AR documents. (2) Synthetic QAs also teach domain-specific factual knowledge: the best-performing recipe mixes both synthetic QAs and synthetic AR documents from QuaLITY, outperforming the mixture of LongHealth synthetic QAs and QuaLITY synthetic AR documents. This suggests that target-domain synthetic QAs provide not only transferable behavior but also domain-specific knowledge.

\subsection{Focal Rewriting: Improving the topic diversity of synthetic documents}

In Section~\ref{sec:prelim}, we show that scaling synthetic documents yields diminishing returns, which we hypothesize is due to limited diversity in the generated data. In particular, WRAP and AR documents produce \textit{stylistically} diverse documents, but the generated documents often cover highly similar \textit{topics}. This is because these approaches do not explicitly condition the LM on specific topics; instead, the model implicitly decides what to focus on, leading to repeated or overlapping topics across generations. In contrast, EG generates documents covering more diverse \textit{topics} by explicitly conditioning on different entities, but the resulting documents tend to be \textit{stylistically} similar. We suspect that this limited diversity degrades performance when synthetic data is scaled extensively, for example, when the number of synthetic tokens exceeds that of the original documents by more than 100 times.

\begin{figure}[t!]
\centering
\includegraphics[width=0.7\linewidth]{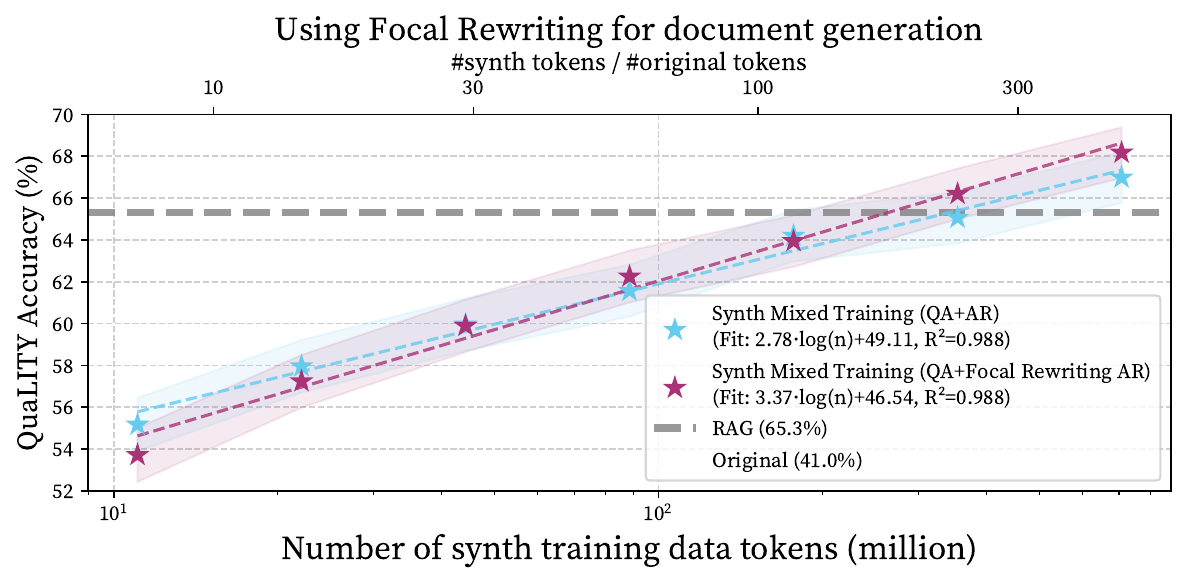}
\caption{\textbf{Scaling synthetic document generation with Focal Rewriting.} We apply Focal Rewriting to AR when generating synthetic documents. Synthetic documents generated with Focal Rewriting (purple line) exhibit better scaling behavior than those generated without it (skyblue line), as shown by the steeper slope of the fitted scaling curve. Data are all generated using 70B model.} 
\label{fig:mix-ours}

  \vspace{-1em}
\end{figure}

Based on this observation, we introduce \textbf{Focal Rewriting} which can diversify both the content and the style of generated documents. When rewriting documents with AR (or WRAP), we explicitly condition generation on a specific question, asking for a document that would be useful for answering that query. This technique can be implemented by simply adding the clause \texttt{``Focus on the question \{\{query\}\}''} in the generation instruction (see Appendix~\ref{sec:appendix-gen} for the prompts). We use the questions generated by LM as in Section~\ref{sec:prelim}. As shown in Appendix~\ref{sec:appendix-more}, when we apply Focal Rewriting to AR, this produces documents with greater lexical and semantic diversity. 

Figure~\ref{fig:mix-ours} shows the results of scaling synthetic data using AR with Focal Rewriting. We find that when doing Synthetic Mixed Training, accuracy follows a log-linear relationship with the number of synthetic tokens. Accordingly, we fit a log-linear curve and plot it alongside the empirical results. The results show that Focal Rewriting yields a steeper log-linear scaling curve and achieves higher accuracy when data is scaled extensively (beyond 175M tokens; $100\times$ more tokens than the original data).

\subsection{Testing on different model and benchmarks}\label{subsec:gen}

\begin{figure}[t!]
  \centering
  \begin{subfigure}[t]{0.49\columnwidth}
    \centering
    \includegraphics[width=\linewidth]{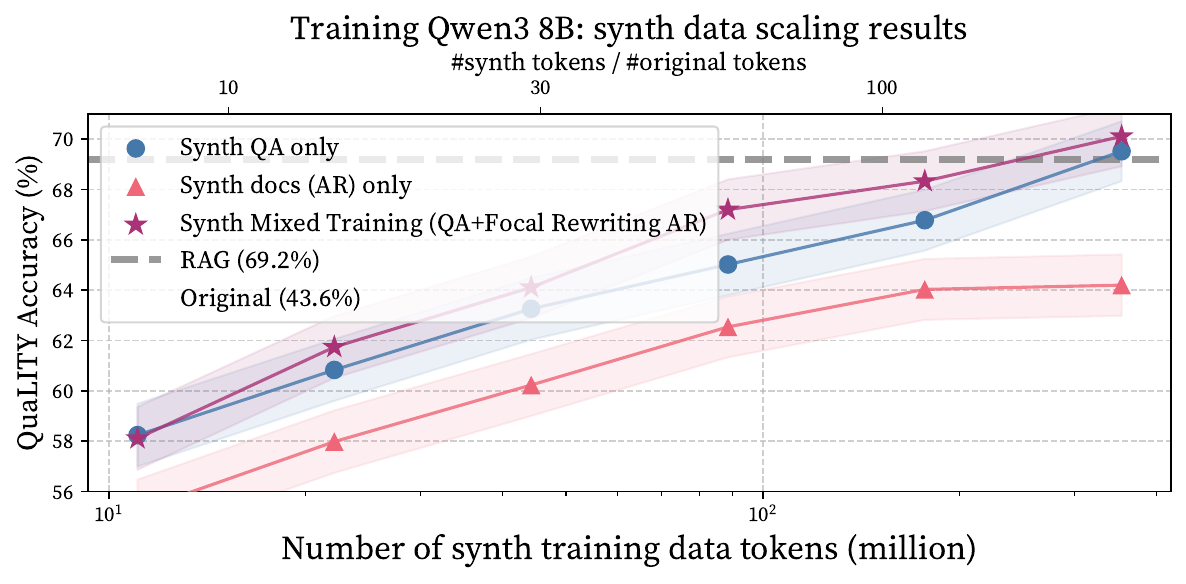}
    \label{fig:quality-qwen3}
  \end{subfigure}\hfill
  \begin{subfigure}[t]{0.49\columnwidth}
    \centering
    \includegraphics[width=\linewidth]{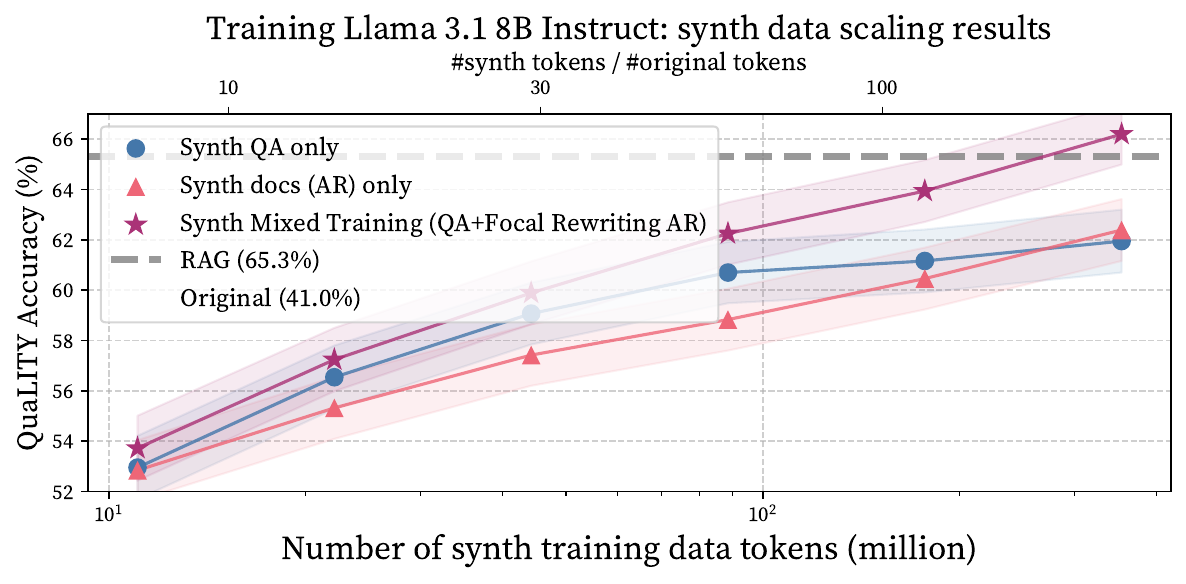}
    \label{fig:quality-llama3}
  \end{subfigure}
  \begin{subfigure}[t]{0.49\columnwidth}
    \centering
    \includegraphics[width=\linewidth]{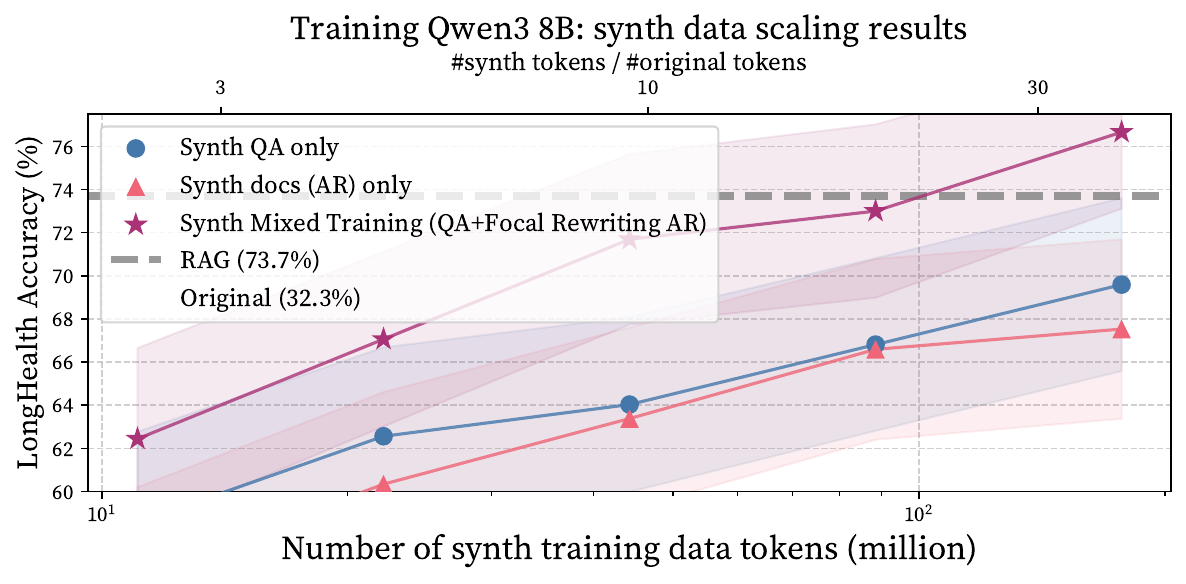}
    \label{fig:longhealth-qwen3}
  \end{subfigure}\hfill
  \begin{subfigure}[t]{0.49\columnwidth}
    \centering
    \includegraphics[width=\linewidth]{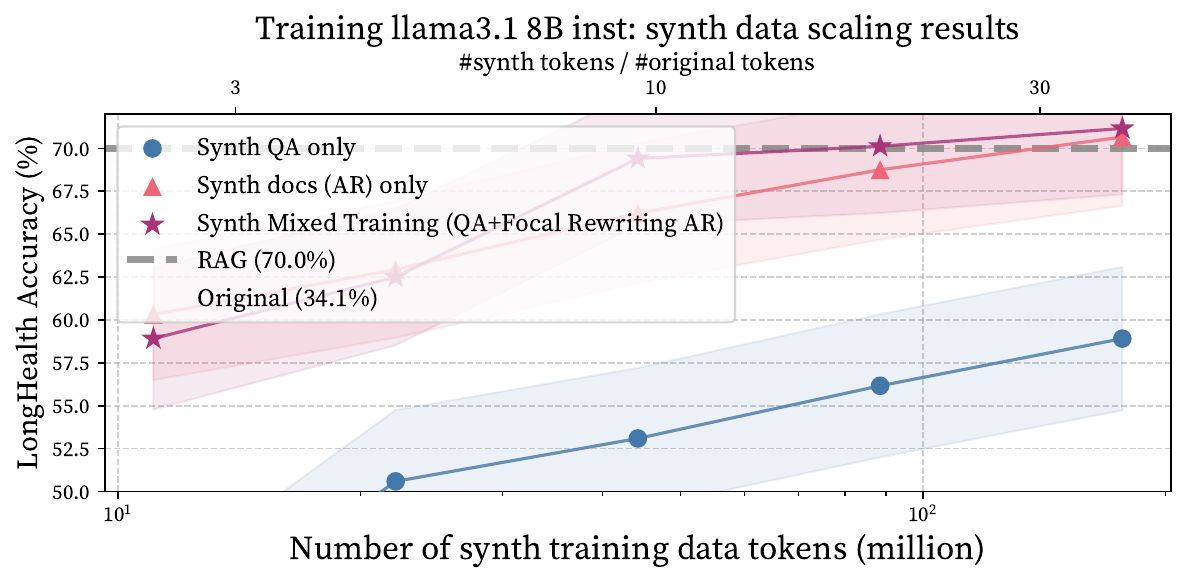}
    \label{fig:longhealth-llama3}
  \end{subfigure}
  \begin{subfigure}[t]{0.49\columnwidth}
    \centering
    \includegraphics[width=\linewidth]{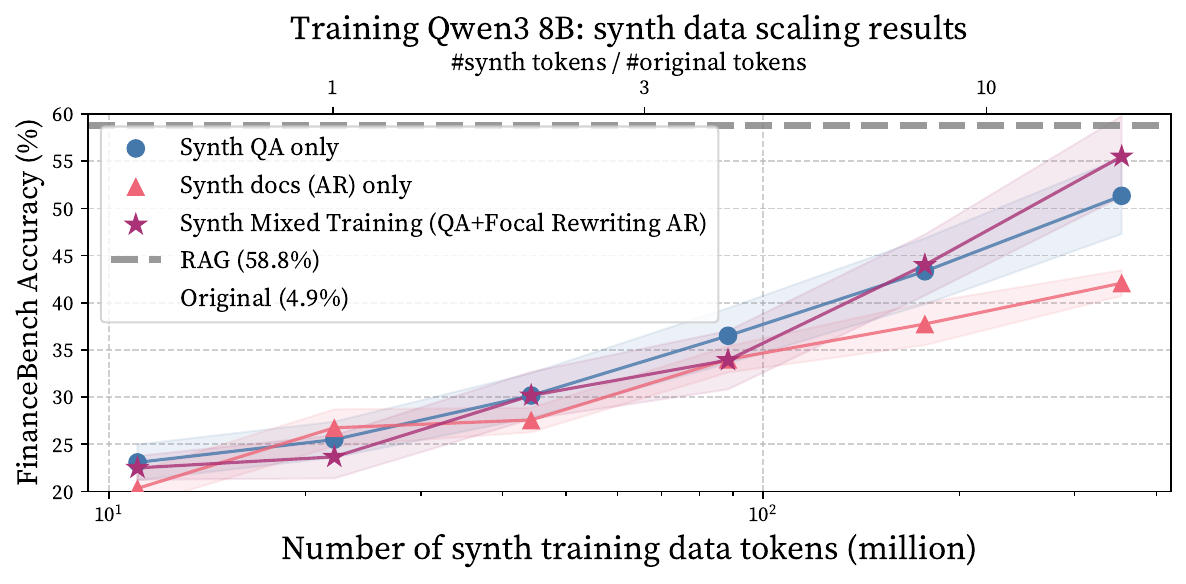}
    \label{fig:financebench-qwen3}
  \end{subfigure}\hfill
  \begin{subfigure}[t]{0.49\columnwidth}
    \centering
    \includegraphics[width=\linewidth]{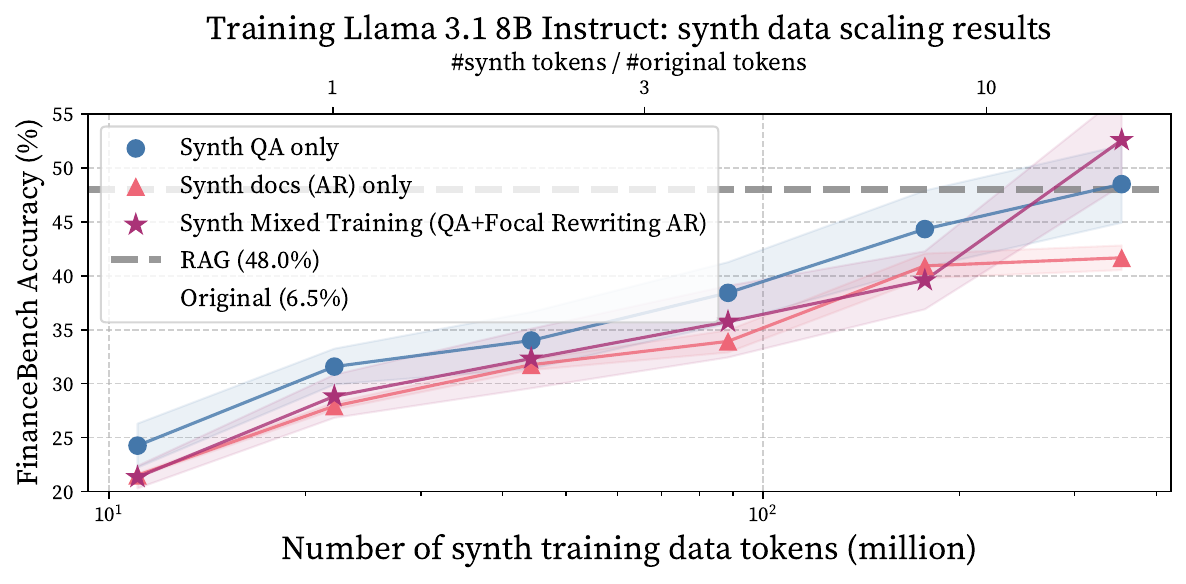}
    \label{fig:financebench-llama3}
  \end{subfigure}
  
  \vspace{-1em}
  \caption{Training Qwen3 8B and Llama 3.1 8B Instruct using synthetic data (generated with 70B model) across three benchmarks (QuaLITY, LongHealth, FinanceBench). Using our recipe (purple line) enables beating RAG on 5/6 of the setups.
  }
  \label{fig:all-benchmarks}
  
  \vspace{-1em}
\end{figure}

\begin{table}
\centering
\small
\begin{tabular}{lrrrrr}
\toprule
Dataset & \#docs & \#avg tokens/docs & \#eval & Eval type & Domain\\
\midrule
QuaLITY & 265 & 6K & 4609 & MCQA & Fictional stories \\ 
LongHealth & 400 & 12K & 400 & MCQA & Medical\\
FinanceBench & 1367 & 16K & 150 & Free-form & Finance\\
\bottomrule
\end{tabular}
\caption{\textbf{Dataset statistics.} \texttt{$\#$docs} indicates the number of source documents used for data generation, and \texttt{$\#$avg tokens/docs} indicates the average number of tokens per source document. \texttt{$\#$eval} indicates the number of QA sets used in evaluation. For FinanceBench, we use Qwen3-14B to judge the model-generated answer using gold answer as a reference.}
\label{table:data_stats}
\vspace{-1em}
\end{table}

We additionally verify our recipe on another base model (Qwen3 8B;~\citet{yang2025qwen3}) and two additional benchmarks (LongHealth;~\citet{adams2025longhealth} and FinanceBench;~\citet{islam2023financebench}) that require learning new knowledge. Table~\ref{table:data_stats} shows the key statistics of the datasets. In particular, FinanceBench provides source documents in PDF format, so we use \texttt{olmOCR-2-7B-1025}~\citep{poznanski2025olmocr} to preprocess them into Markdown and use them as source documents.

Figure~\ref{fig:all-benchmarks} shows the results. On QuaLITY and LongHealth, our training recipe enables the models to outperform RAG, and on FinanceBench, our methods allow the Llama 8B model to outperform RAG. On average, our method gives 2.6\% relative accuracy gain compared to RAG. Our method also outperforms training recipes that scale only synthetic QAs or synthetic AR documents.

\begin{figure}[t!]
\centering
\includegraphics[width=0.9\linewidth]{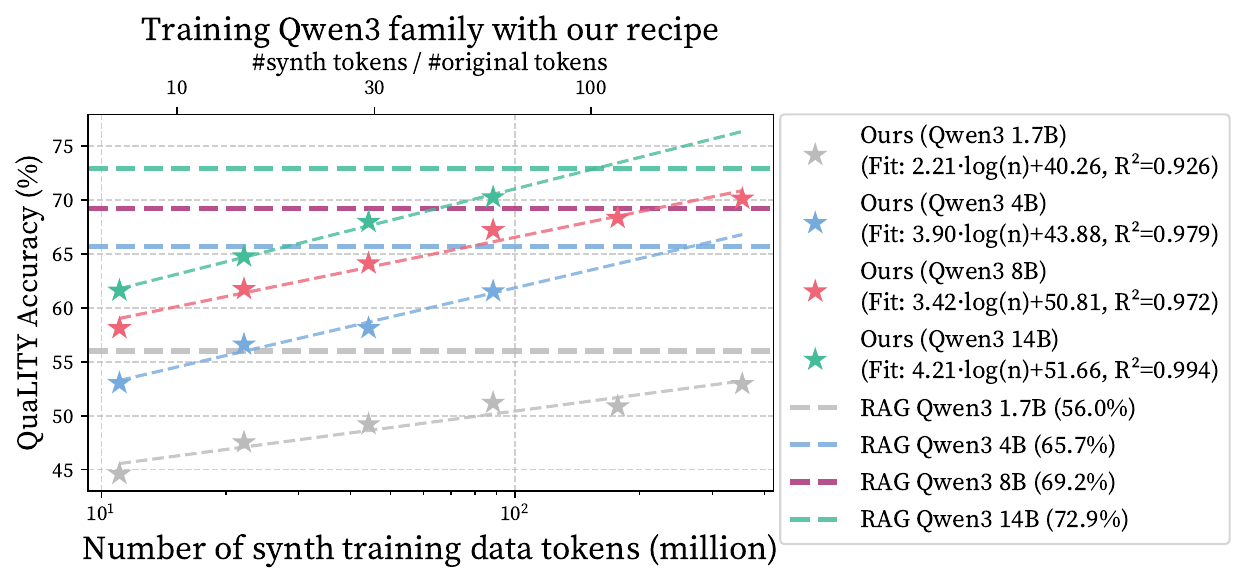}
\caption{\textbf{Training different-sized models from the Qwen3 family on the synthetic QuaLITY dataset.} We train using our best recipe: Mixed Synthetic Training with synthetic QAs and Focal Rewriting AR documents, using 70B generator. All models exhibit log-linear scaling behavior, and larger models can match RAG performance with fewer synthetic tokens. Based on the fitted curves, we observe that the 14B model requires 102$\times$, the 8B model 142$\times$, the 4B model 177$\times$, and the 1.7B model 813$\times$ more synthetic tokens than the original token count (shown as \texttt{\#synth tokens/\#original tokens)}. 
}
\vspace{-1em}
\label{fig:qwen-family}
\end{figure}

\subsection{Training larger models is more synthetic token efficient}

We train four Qwen3 models of different sizes (1.7B, 4B, 8B, 14B) to study how model size affects scaling behavior. We keep all configurations (including the learning rate) the same from the previous experiments. Figure~\ref{fig:qwen-family} shows that these models all show log-linear scaling when tokens are scaled up to 88M, and larger models are more synthetic token efficient, achieving RAG-level performance with fewer synthetic tokens according to the fitted log-linear curve. For instance, the 14B model requires 102$\times$ more synthetic tokens than the original data, whereas the 1.7B model requires 813$\times$ more. This result is intuitive, as larger models have greater capacity to store knowledge~\citep{morris2025much}.

\subsection{Our training recipe enhances RAG} We test whether our trained model can be improved further when paired with retrieval augmentation. As shown in Table~\ref{tab:rag_complements_training}, this shows clear improvement when using RAG with our trained model and shows that it significantly outperforms the vanilla RAG baselines. On average, our trained models combined with RAG provides a 9.1\% relative gain when compared to RAG. This suggests that domain-specific training with synthetic data augmentation can be helpful even when RAG is used.

\begin{table}[t]
\centering
\small
\setlength{\tabcolsep}{5pt}
\begin{tabular}{llrrrr}
\toprule
\textbf{Benchmark} & \textbf{Model} & \textbf{Ours} & \textbf{Ours + RAG} & \textbf{Vanilla RAG} & \textbf{$\Delta$} \\
\midrule
\multirow{2}{*}{QuaLITY}
  & Llama & 68.2\% & 69.7\% & 65.3\% & +4.4 \\
  & Qwen  & 70.1\% & 73.6\% & 69.2\% & +4.4 \\
\midrule
\multirow{2}{*}{LongHealth}
  & Llama & 71.2\% & 80.3\% & 70.0\% & +10.3 \\
  & Qwen  & 76.7\% & 82.3\% & 73.7\% & +8.6 \\
\midrule
\multirow{2}{*}{FinanceBench}
  & Llama & 52.6\% & 54.3\% & 48.0\% & +6.3 \\
  & Qwen  & 55.5\% & 60.2\% & 58.8\% & +1.4 \\
\bottomrule
\end{tabular}
\caption{\textbf{Our training complements retrieval augmentation.} ``Ours'' denotes the trained model without RAG, ``Ours + RAG'' denotes the same trained model with RAG, and ``Vanilla RAG'' denotes the RAG baseline. $\Delta$ denotes the absolute improvement of ``Ours + RAG'' over ``Vanilla RAG''. Averaged across all settings, our training improves over Vanilla RAG by 5.9 points.}
\label{tab:rag_complements_training}
\vspace{-1em}
\end{table}

\section{Related Work}
\paragraph{Training with synthetic data.} 
Training language models with synthetic data at scale has become an important practice~\citep{blakeman2025nvidia, yang2025qwen3, abdin2024phi}. At the pre-training stage, language models are often used to rewrite original data, such as web text. For example, \citet{maini2024rephrasing} use language models to rephrase original documents, while \citet{nguyenrecycling} use language models to reason before rephrasing documents, resulting in higher-quality rewrites. More recently, \citet{yang2025synthetic} propose to train a language model to generate new documents conditioned on an input document. 

In continued pre-training settings, where language models are further trained on domain-specific data after pre-training~\citep{gururangan2020don}, more aggressive forms of data augmentation are often used because these settings are typically data-constrained and do not provide enough data for scaling~\citep{muennighoff2023scaling,kim2025pre}. For these settings, improving the diversity of generated data is important: to this end, \citet{yangsynthetic} uses language models to extract core entities about the documents, and then generate synthetic documents that describe relations between entities in the original documents, thereby improving the diversity of the synthetic data. They empirically show that accuracy improves in a log-linear trend as the number of synthetic data tokens increases. Similarly, \citet{lin2025learning} use language models to diversify rewriting strategies. In a more domain-specific direction, \citet{ruan2025reasoning} use reasoning traces produced by language models to capture the underlying thought processes related to the original documents, and show that these traces are helpful for continued training in math.

\paragraph{Analyzing knowledge of language models.} How language models acquire knowledge during training and use that knowledge when performing downstream tasks is still not fully understood. To help understand this, \citet{allen2024physics} and \citet{allen2023physics} conduct systematic studies on small language models and show that both storing knowledge in the parameters and learning how to use that knowledge are important. More recently, \citet{calderon2026empty} argue that even frontier models are bottlenecked more by knowledge recall than by knowledge storage, and \citet{gekhman2026thinkingrecallreasoningunlocks, ma2026improving} show that reasoning can improve fact recall, indicating that teaching language models how to use the facts learned during training is important. We believe that the success of Synthetic Mixed Training aligns well with these findings, as synthetic QAs may help models learn how to use knowledge, highlighting the importance of knowledge use beyond mere storage in language models.

\section{Conclusion}

We study how to make synthetic data scale more effectively for knowledge learning in data-constrained domains. Our results show that simply increasing the amount of synthetic data or using a stronger generator is not sufficient: existing methods exhibit diminishing returns and still underperform RAG. Based on the observation that synthetic QAs and documents have different scaling properties, we introduce Synthetic Mixed Training, which combines synthetic QAs and documents to leverage their complementary training signals and achieve the best of both worlds. We further introduce Focal Rewriting, which improves the diversity of generated documents and leads to an even steeper scaling trend. Our methods generalize well across a range of settings, and the trained models are also complementary to RAG.

\section{Acknowledgments}

We thank Suhong Moon and Sehoon Kim for their valuable feedback and support throughout this work. We also acknowledge Upstage, and Vessl for their compute support for this work, and thank Singapore DSO and the Institute of Information \& Communications Technology Planning \& Evaluation (IITP) grant funded by the Korean MSIT (No. RS-2024-00457882, National AI Research Lab Project) for supporting this work.

\bibliographystyle{plainnat}
\bibliography{reference}

@inproceedings{yangsynthetic,
  title={Synthetic continued pretraining},
  author={Yang, Zitong and Band, Neil and Li, Shuangping and Candes, Emmanuel and Hashimoto, Tatsunori},
  booktitle={The Thirteenth International Conference on Learning Representations},
  year={2025}
}

@inproceedings{maini2024rephrasing,
  title={Rephrasing the web: A recipe for compute and data-efficient language modeling},
  author={Maini, Pratyush and Seto, Skyler and Bai, Richard and Grangier, David and Zhang, Yizhe and Jaitly, Navdeep},
  booktitle={Proceedings of the 62nd Annual Meeting of the Association for Computational Linguistics (Volume 1: Long Papers)},
  pages={14044--14072},
  year={2024}
}

@article{lin2025learning,
  title={Learning facts at scale with active reading},
  author={Lin, Jessy and Berges, Vincent-Pierre and Chen, Xilun and Yih, Wen-Tau and Ghosh, Gargi and O{\u{g}}uz, Barlas},
  journal={arXiv preprint arXiv:2508.09494},
  year={2025}
}

@article{maini2025beyondweb,
  title={Beyondweb: Lessons from scaling synthetic data for trillion-scale pretraining},
  author={Maini, Pratyush and Dorna, Vineeth and Doshi, Parth and Carranza, Aldo and Pan, Fan and Urbanek, Jack and Burstein, Paul and Fang, Alex and Deng, Alvin and Abbas, Amro and others},
  journal={arXiv preprint arXiv:2508.10975},
  year={2025}
}

@article{abdin2024phi,
  title={Phi-4 technical report},
  author={Abdin, Marah and Aneja, Jyoti and Behl, Harkirat and Bubeck, S{\'e}bastien and Eldan, Ronen and Gunasekar, Suriya and Harrison, Michael and Hewett, Russell J and Javaheripi, Mojan and Kauffmann, Piero and others},
  journal={arXiv preprint arXiv:2412.08905},
  year={2024}
}

@article{muennighoff2023scaling,
  title={Scaling data-constrained language models},
  author={Muennighoff, Niklas and Rush, Alexander and Barak, Boaz and Le Scao, Teven and Tazi, Nouamane and Piktus, Aleksandra and Pyysalo, Sampo and Wolf, Thomas and Raffel, Colin A},
  journal={Advances in Neural Information Processing Systems},
  volume={36},
  pages={50358--50376},
  year={2023}
}

@article{kim2025pre,
  title={Pre-training under infinite compute},
  author={Kim, Konwoo and Kotha, Suhas and Liang, Percy and Hashimoto, Tatsunori},
  journal={arXiv preprint arXiv:2509.14786},
  year={2025}
}

@article{guha2025openthoughts,
  title={Openthoughts: Data recipes for reasoning models},
  author={Guha, Etash and Marten, Ryan and Keh, Sedrick and Raoof, Negin and Smyrnis, Georgios and Bansal, Hritik and Nezhurina, Marianna and Mercat, Jean and Vu, Trung and Sprague, Zayne and others},
  journal={arXiv preprint arXiv:2506.04178},
  year={2025}
}

@article{eyuboglu2025cartridges,
  title={Cartridges: Lightweight and general-purpose long context representations via self-study},
  author={Eyuboglu, Sabri and Ehrlich, Ryan and Arora, Simran and Guha, Neel and Zinsley, Dylan and Liu, Emily and Tennien, Will and Rudra, Atri and Zou, James and Mirhoseini, Azalia and others},
  journal={arXiv preprint arXiv:2506.06266},
  year={2025}
}

@inproceedings{cacciatraining,
  title={Training Plug-and-Play Knowledge Modules with Deep Context Distillation},
  author={Caccia, Lucas and Ansell, Alan and Ponti, Edoardo and Vuli{\'c}, Ivan and Sordoni, Alessandro},
  booktitle={Second Conference on Language Modeling},
  year={2025}
}

@article{lewis2020retrieval,
  title={Retrieval-augmented generation for knowledge-intensive nlp tasks},
  author={Lewis, Patrick and Perez, Ethan and Piktus, Aleksandra and Petroni, Fabio and Karpukhin, Vladimir and Goyal, Naman and K{\"u}ttler, Heinrich and Lewis, Mike and Yih, Wen-tau and Rockt{\"a}schel, Tim and others},
  journal={Advances in neural information processing systems},
  volume={33},
  pages={9459--9474},
  year={2020}
}

@article{adams2025longhealth,
  title={Longhealth: A question answering benchmark with long clinical documents},
  author={Adams, Lisa and Busch, Felix and Han, Tianyu and Excoffier, Jean-Baptiste and Ortala, Matthieu and L{\"o}ser, Alexander and Aerts, Hugo JWL and Kather, Jakob Nikolas and Truhn, Daniel and Bressem, Keno},
  journal={Journal of Healthcare Informatics Research},
  volume={9},
  number={3},
  pages={280--296},
  year={2025},
  publisher={Springer}
}

@article{islam2023financebench,
  title={Financebench: A new benchmark for financial question answering},
  author={Islam, Pranab and Kannappan, Anand and Kiela, Douwe and Qian, Rebecca and Scherrer, Nino and Vidgen, Bertie},
  journal={arXiv preprint arXiv:2311.11944},
  year={2023}
}

@article{penedo2024fineweb,
  title={The fineweb datasets: Decanting the web for the finest text data at scale},
  author={Penedo, Guilherme and Kydl{\'\i}{\v{c}}ek, Hynek and Lozhkov, Anton and Mitchell, Margaret and Raffel, Colin A and Von Werra, Leandro and Wolf, Thomas and others},
  journal={Advances in Neural Information Processing Systems},
  volume={37},
  pages={30811--30849},
  year={2024}
}

@inproceedings{pang-etal-2022-quality,
    title = "{Q}u{ALITY}: Question Answering with Long Input Texts, Yes!",
    author = "Pang, Richard Yuanzhe  and
      Parrish, Alicia  and
      Joshi, Nitish  and
      Nangia, Nikita  and
      Phang, Jason  and
      Chen, Angelica  and
      Padmakumar, Vishakh  and
      Ma, Johnny  and
      Thompson, Jana  and
      He, He  and
      Bowman, Samuel",
    booktitle = "Proceedings of the 2022 Conference of the North American Chapter of the Association for Computational Linguistics: Human Language Technologies",
    month = jul,
    year = "2022",
    address = "Seattle, United States",
    publisher = "Association for Computational Linguistics",
    url = "https://aclanthology.org/2022.naacl-main.391",
    pages = "5336--5358",
}

@article{grattafiori2024llama,
  title={The llama 3 herd of models},
  author={Grattafiori, Aaron and Dubey, Abhimanyu and Jauhri, Abhinav and Pandey, Abhinav and Kadian, Abhishek and Al-Dahle, Ahmad and Letman, Aiesha and Mathur, Akhil and Schelten, Alan and Vaughan, Alex and others},
  journal={arXiv preprint arXiv:2407.21783},
  year={2024}
}

@inproceedings{allen2024physics,
  title={Physics of Language Models: Part 3.1, Knowledge Storage and Extraction},
  author={Allen-Zhu, Zeyuan and Li, Yuanzhi},
  booktitle={International Conference on Machine Learning},
  pages={1067--1077},
  year={2024},
  organization={PMLR}
}

@inproceedings{allen2023physics,
  title={Physics of Language Models: Part 3.2, Knowledge Manipulation},
  author={Allen-Zhu, Zeyuan and Li, Yuanzhi},
  booktitle={International Conference on Learning Representations},
  year={2025}
}

@article{yang2025qwen3,
  title={Qwen3 technical report},
  author={Yang, An and Li, Anfeng and Yang, Baosong and Zhang, Beichen and Hui, Binyuan and Zheng, Bo and Yu, Bowen and Gao, Chang and Huang, Chengen and Lv, Chenxu and others},
  journal={arXiv preprint arXiv:2505.09388},
  year={2025}
}

@inproceedings{nguyenrecycling,
  title={Recycling the Web: A Method to Enhance Pre-training Data Quality and Quantity for Language Models},
  author={Nguyen, Thao and Li, Yang and Golovneva, Olga and Zettlemoyer, Luke and Oh, Sewoong and Schmidt, Ludwig and Li, Xian},
  booktitle={Second Conference on Language Modeling},
  year={2025}
}

@inproceedings{kwon2023efficient,
  title={Efficient memory management for large language model serving with pagedattention},
  author={Kwon, Woosuk and Li, Zhuohan and Zhuang, Siyuan and Sheng, Ying and Zheng, Lianmin and Yu, Cody Hao and Gonzalez, Joseph and Zhang, Hao and Stoica, Ion},
  booktitle={Proceedings of the 29th symposium on operating systems principles},
  pages={611--626},
  year={2023}
}

@inproceedings{jungprismatic,
  title={Prismatic Synthesis: Gradient-based Data Diversification Boosts Generalization in LLM Reasoning},
  author={Jung, Jaehun and Han, Seungju and Lu, Ximing and Hallinan, Skyler and Acuna, David and Prabhumoye, Shrimai and Patwary, Mostofa and Shoeybi, Mohammad and Catanzaro, Bryan and Choi, Yejin},
  booktitle={The Thirty-ninth Annual Conference on Neural Information Processing Systems},
  year={2025}
}

@article{johnson1984extensions,
  title={Extensions of Lipschitz mappings into a Hilbert space},
  author={Johnson, William B and Lindenstrauss, Joram and others},
  journal={Contemporary mathematics},
  volume={26},
  number={189-206},
  pages={1},
  year={1984}
}

@article{poznanski2025olmocr,
  title={olmocr 2: Unit test rewards for document ocr},
  author={Poznanski, Jake and Soldaini, Luca and Lo, Kyle},
  journal={arXiv preprint arXiv:2510.19817},
  year={2025}
}

@article{blakeman2025nvidia,
  title={NVIDIA Nemotron 3: Efficient and Open Intelligence},
  author={Blakeman, Aaron and Grattafiori, Aaron and Basant, Aarti and Gupta, Abhibha and Khattar, Abhinav and Renduchintala, Adi and Vavre, Aditya and Shukla, Akanksha and Bercovich, Akhiad and Ficek, Aleksander and others},
  journal={arXiv preprint arXiv:2512.20856},
  year={2025}
}

@article{ruan2025reasoning,
  title={Reasoning to learn from latent thoughts},
  author={Ruan, Yangjun and Band, Neil and Maddison, Chris J and Hashimoto, Tatsunori},
  journal={arXiv preprint arXiv:2503.18866},
  year={2025}
}

@article{yang2025synthetic,
  title={Synthetic bootstrapped pretraining},
  author={Yang, Zitong and Zhang, Aonan and Liu, Hong and Hashimoto, Tatsunori and Cand{\`e}s, Emmanuel and Wang, Chong and Pang, Ruoming},
  journal={arXiv preprint arXiv:2509.15248},
  year={2025}
}

@article{zweiger2026fast,
  title={Fast KV Compaction via Attention Matching},
  author={Zweiger, Adam and Fu, Xinghong and Guo, Han and Kim, Yoon},
  journal={arXiv preprint arXiv:2602.16284},
  year={2026}
}

@inproceedings{hulora,
  title={LoRA: Low-Rank Adaptation of Large Language Models},
  author={Hu, Edward J and Wallis, Phillip and Allen-Zhu, Zeyuan and Li, Yuanzhi and Wang, Shean and Wang, Lu and Chen, Weizhu and others},
  booktitle={International Conference on Learning Representations},
  year={2022}
}

@article{bidermanlora,
  title={LoRA Learns Less and Forgets Less},
  author={Biderman, Dan and Portes, Jacob and Ortiz, Jose Javier Gonzalez and Paul, Mansheej and Greengard, Philip and Jennings, Connor and King, Daniel and Havens, Sam and Chiley, Vitaliy and Frankle, Jonathan and others},
  journal={Transactions on Machine Learning Research},
  year={2025}
}

@inproceedings{geva2021transformer,
  title={Transformer feed-forward layers are key-value memories},
  author={Geva, Mor and Schuster, Roei and Berant, Jonathan and Levy, Omer},
  booktitle={Proceedings of the 2021 Conference on Empirical Methods in Natural Language Processing},
  pages={5484--5495},
  year={2021}
}

@article{snell2022learning,
  title={Learning by distilling context},
  author={Snell, Charlie and Klein, Dan and Zhong, Ruiqi},
  journal={arXiv preprint arXiv:2209.15189},
  year={2022}
}

@article{calderon2026empty,
  title={Empty Shelves or Lost Keys? Recall Is the Bottleneck for Parametric Factuality},
  author={Calderon, Nitay and Ben-David, Eyal and Gekhman, Zorik and Ofek, Eran and Yona, Gal},
  journal={arXiv preprint arXiv:2602.14080},
  year={2026}
}

@misc{niklaus2026_the_synthetic_data_playbook_generating_trillions_of_the_finest_tokens,
  title={The Synthetic Data Playbook: Generating Trillions of the Finest Tokens},
  author={Joel Niklaus and Guilherme Penedo and Hynek Kydlicek and Elie Bakouch and Lewis Tunstall and Ed Beeching and Thibaud Frere and Colin Raffel and Leandro von Werra and Thomas Wolf},
  year={2026}, 
}

@misc{gekhman2026thinkingrecallreasoningunlocks,
      title={Thinking to Recall: How Reasoning Unlocks Parametric Knowledge in LLMs}, 
      author={Zorik Gekhman and Roee Aharoni and Eran Ofek and Mor Geva and Roi Reichart and Jonathan Herzig},
      year={2026},
      eprint={2603.09906},
      archivePrefix={arXiv},
      primaryClass={cs.CL},
      url={https://arxiv.org/abs/2603.09906}, 
}

@article{friedmanvendi,
  title={The Vendi Score: A Diversity Evaluation Metric for Machine Learning},
  author={Friedman, Dan and Dieng, Adji Bousso},
  journal={Transactions on Machine Learning Research},
  year={2023}
}

@article{morris2025much,
  title={How much do language models memorize?},
  author={Morris, John X and Sitawarin, Chawin and Guo, Chuan and Kokhlikyan, Narine and Suh, G Edward and Rush, Alexander M and Chaudhuri, Kamalika and Mahloujifar, Saeed},
  journal={arXiv preprint arXiv:2505.24832},
  year={2025}
}

@inproceedings{kang2025demystifying,
  title={Demystifying synthetic data in llm pre-training: A systematic study of scaling laws, benefits, and pitfalls},
  author={Kang, Feiyang and Ardalani, Newsha and Kuchnik, Michael and Emad, Youssef and Elhoushi, Mostafa and Sengupta, Shubhabrata and Li, Shang-Wen and Raghavendra, Ramya and Jia, Ruoxi and Wu, Carole-Jean},
  booktitle={Proceedings of the 2025 Conference on Empirical Methods in Natural Language Processing},
  pages={10750--10769},
  year={2025}
}

@article{lampinen2025latent,
  title={Latent learning: episodic memory complements parametric learning by enabling flexible reuse of experiences},
  author={Lampinen, Andrew Kyle and Engelcke, Martin and Li, Yuxuan and Chaudhry, Arslan and McClelland, James L},
  journal={arXiv preprint arXiv:2509.16189},
  year={2025}
}

@inproceedings{gururangan2020don,
  title={Don’t stop pretraining: Adapt language models to domains and tasks},
  author={Gururangan, Suchin and Marasovi{\'c}, Ana and Swayamdipta, Swabha and Lo, Kyle and Beltagy, Iz and Downey, Doug and Smith, Noah A},
  booktitle={Proceedings of the 58th annual meeting of the association for computational linguistics},
  pages={8342--8360},
  year={2020}
}

@article{ma2026improving,
  title={Improving Parametric Knowledge Access in Reasoning Language Models},
  author={Ma, Melody and Hewitt, John},
  journal={arXiv preprint arXiv:2602.22193},
  year={2026}
}

@article{gao2025metadata,
  title={Metadata conditioning accelerates language model pre-training},
  author={Gao, Tianyu and Wettig, Alexander and He, Luxi and Dong, Yihe and Malladi, Sadhika and Chen, Danqi},
  journal={arXiv preprint arXiv:2501.01956},
  year={2025}
}

@article{loshchilov2017decoupled,
  title={Decoupled weight decay regularization},
  author={Loshchilov, Ilya and Hutter, Frank},
  journal={arXiv preprint arXiv:1711.05101},
  year={2017}
}

@article{zhao2023pytorch,
  title={Pytorch fsdp: experiences on scaling fully sharded data parallel},
  author={Zhao, Yanli and Gu, Andrew and Varma, Rohan and Luo, Liang and Huang, Chien-Chin and Xu, Min and Wright, Less and Shojanazeri, Hamid and Ott, Myle and Shleifer, Sam and others},
  journal={arXiv preprint arXiv:2304.11277},
  year={2023}
}

@article{kotha2026replaying,
  title={Replaying pre-training data improves fine-tuning},
  author={Kotha, Suhas and Liang, Percy},
  journal={arXiv preprint arXiv:2603.04964},
  year={2026}
}

@article{liu2025midtraining,
  title={Midtraining Bridges Pretraining and Posttraining Distributions},
  author={Liu, Emmy and Neubig, Graham and Xiong, Chenyan},
  journal={arXiv preprint arXiv:2510.14865},
  year={2025}
}

@article{lin2025continual,
  title={Continual learning via sparse memory finetuning},
  author={Lin, Jessy and Zettlemoyer, Luke and Ghosh, Gargi and Yih, Wen-Tau and Markosyan, Aram and Berges, Vincent-Pierre and O{\u{g}}uz, Barlas},
  journal={arXiv preprint arXiv:2510.15103},
  year={2025}
}

@article{berges2024memory,
  title={Memory layers at scale},
  author={Berges, Vincent-Pierre and O{\u{g}}uz, Barlas and Haziza, Daniel and Yih, Wen-tau and Zettlemoyer, Luke and Ghosh, Gargi},
  journal={arXiv preprint arXiv:2412.09764},
  year={2024}
}

@article{he2024mixture,
  title={Mixture of a million experts},
  author={He, Xu Owen},
  journal={arXiv preprint arXiv:2407.04153},
  year={2024}
}

@article{shazeer2017outrageously,
  title={Outrageously large neural networks: The sparsely-gated mixture-of-experts layer},
  author={Shazeer, Noam and Mirhoseini, Azalia and Maziarz, Krzysztof and Davis, Andy and Le, Quoc and Hinton, Geoffrey and Dean, Jeff},
  journal={arXiv preprint arXiv:1701.06538},
  year={2017}
}

@inproceedings{agarwal2024policy,
  title={On-policy distillation of language models: Learning from self-generated mistakes},
  author={Agarwal, Rishabh and Vieillard, Nino and Zhou, Yongchao and Stanczyk, Piotr and Garea, Sabela Ramos and Geist, Matthieu and Bachem, Olivier},
  booktitle={The twelfth international conference on learning representations},
  year={2024}
}

@article{lu2025onpolicydistillation,
  author = {Kevin Lu and Thinking Machines Lab},
  title = {On-Policy Distillation},
  journal = {Thinking Machines Lab: Connectionism},
  year = {2025},
  note = {https://thinkingmachines.ai/blog/on-policy-distillation},
  doi = {10.64434/tml.20251026},
}

@article{kaplan2020scaling,
  title={Scaling laws for neural language models},
  author={Kaplan, Jared and McCandlish, Sam and Henighan, Tom and Brown, Tom B and Chess, Benjamin and Child, Rewon and Gray, Scott and Radford, Alec and Wu, Jeffrey and Amodei, Dario},
  journal={arXiv preprint arXiv:2001.08361},
  year={2020}
}

@inproceedings{soudani2024fine,
  title={Fine tuning vs. retrieval augmented generation for less popular knowledge},
  author={Soudani, Heydar and Kanoulas, Evangelos and Hasibi, Faegheh},
  booktitle={Proceedings of the 2024 Annual International ACM SIGIR Conference on Research and Development in Information Retrieval in the Asia Pacific Region},
  pages={12--22},
  year={2024}
}

@inproceedings{ovadia2024fine,
  title={Fine-tuning or retrieval? comparing knowledge injection in llms},
  author={Ovadia, Oded and Brief, Menachem and Mishaeli, Moshik and Elisha, Oren},
  booktitle={Proceedings of the 2024 conference on empirical methods in natural language processing},
  pages={237--250},
  year={2024}
}

\newpage
\appendix

\section{Limitations} Due to compute limitations, our study focuses on small-scale models up to 8B parameters, which stand to benefit the most from novel methods for knowledge learning. In addition, although we focus on learning new knowledge, mitigating the forgetting of existing knowledge is also an important problem. We use pretraining data replay~\citep{yangsynthetic, kotha2026replaying} to mitigate this forgetting issue, and we view a deeper treatment of forgetting alongside knowledge acquisition as a promising direction for future work.

\section{Training details}\label{sec:appendix-training}
\paragraph{Hyperparameters.} We train all models on the synthetic data using a fixed set of hyperparameters: a batch size of 16, a sequence length of 2048, two training epochs, and a fixed replay rate of 0.1 from the FineWeb dataset~\citep{penedo2024fineweb}. We use cosine learning rate schedule with warm up ratio of 0.05, use AdamW optimizer~\citep{loshchilov2017decoupled} with weight decay of 0.01, beta1 of 0.9, and beta2 of 0.999. We also use gradient clipping of threshold 1.0 and use FSDP2~\citep{zhao2023pytorch} for model training (used 4 GPUs for training 8B models). The only variation is the method used to generate the synthetic data. For Llama 3.1 8B Instruct, we train models with two learning rates (5e-6 and 1e-5) and report the best result for each configuration. For Qwen3 models (1.7B, 4B, 8B, and 14B), we use two learning rates (1e-5 and 5e-5) and again report the best result for each configuration. To use training compute efficiently, as in pretraining, we pack randomly shuffled data instances into each sequence, separated by the EOD delimiter.

\paragraph{Data formatting: inclusion of metadata is important.} After generating synthetic data, we emphasize that data formatting is also important: metadata about the data should be included (e.g., company name, story title and author name, etc). For example, if the generated data (using Active Reading) is based on FinanceBench, it could follow this format:

\begin{table}[h]
\centering
\begin{tabular}{p{0.9\linewidth}}
\toprule
\texttt{Here's a learning strategy.}\\
\texttt{\{strategy\}}\\
\\
\texttt{Apply this strategy to the document ``\{doc\_name\}'' of \{company\}.}\\
\\
\texttt{Output: \{generated\_text\}} \\
\bottomrule
\end{tabular}
\caption{Example data format for synthetic training data with metadata. This is an example of using FinanceBench matadata.}
\label{tab:data-format}
\end{table}

In our preliminary experiments, we found that omitting metadata leads to lower accuracy after training. This is reasonable, as without metadata, the model struggles to associate knowledge with the correct source. This observation is also consistent with the findings of \citet{allen2023physics} and \citet{gao2025metadata}.

\paragraph{Estimating compute for synthetic training.}
We provide a crude estimate of the amount of compute required for synthetic training. We follow the common approximation for FLOP calculation from \citet{kaplan2020scaling}: $2ND$ for the forward pass and $4ND$ for the backward pass, where $N$ is the number of model parameters and $D$ is the number of data tokens. Under this approximation, when we train a model with $N$ parameters on $D$ synthetic tokens generated by a model with $M$ parameters, the total compute required for synthetic training can be written as $C \approx 2MD + 6ND$. Using this formula, we estimate the computational cost of our most expensive training run: training an 8B model on 700M tokens generated by a 70B model requires approximately $1.316 \times 10^{20}$ FLOPs. Assuming using H100 (1979 TFLOPS) for synthetic training without any other overhead, it requires 18.5 H100 hours to generate synthetic data and train the model on it.

\section{Evaluation details}\label{sec:appendix-eval}
\paragraph{RAG implementation.} We compare the trained model against a RAG~\citep{lewis2020retrieval} that uses the model before synthetic data training. To implement RAG, we use Qwen3-Embedding-8B as the retriever to fetch the top-128 document chunks most relevant to the query. We then apply Qwen3-Reranker-8B to rerank these chunks and select the top-8 as context.

\paragraph{Evaluation hyperparameters.} During evaluation, we use temperature=0.1, top-p=0.95, and a maximum length of 512. We generate eight responses per question (n=8) and report the average accuracy for a more robust evaluation. For model evaluations on the MCQA benchmarks (QuaLITY, LongHealth), we use the prompt in Table~\ref{tab:appendix-prompt}, and for the open-ended generation task (FinanceBench), we use the prompt in Table~\ref{tab:appendix-docqa-prompt}.

\begin{table}[t]
\centering
\begin{tabular}{p{0.95\linewidth}}
\toprule
\texttt{\#\#\# Question} \\
\texttt{\{question\}} \\[0.5em]

\texttt{\#\#\# Choices} \\
\texttt{\{options\}} \\[0.5em]

\texttt{Choose the best answer from the following options after thinking step by step. There is only one correct choice.} \\
\texttt{Your answer format should be like this:} \\
\texttt{Explanation: [your explanation]} \\
\texttt{Answer: [your answer (only one letter, A, B, C, D, or E)]} \\
\bottomrule
\caption{Prompt template used for multiple-choice QA evaluation. This is used for QuaLITY and LongHealth.}
\label{tab:appendix-prompt}
\end{tabular}
\end{table}

\begin{table}[t]
\centering
\begin{tabular}{p{0.95\textwidth}}
\toprule
\texttt{\#\#\# Question} \\
\texttt{\{question\}} \\[0.5em]
\texttt{Answer the question using the document above.} \\
\texttt{Your answer format should be like this:} \\
\texttt{Explanation: [your explanation]} \\
\texttt{Answer: [your answer]} \\
\bottomrule
\caption{Prompt template used for open-ended QA evaluation. This is used for FinanceBench.}
\label{tab:appendix-docqa-prompt}
\end{tabular}
\end{table}

\section{Synthetic Data Generation Details}\label{sec:appendix-gen}

We use vLLM~\citep{kwon2023efficient} for efficient LLM inference during data generation. For QA pair generation, we use the prompt in Table~\ref{tab:qa-prompt}. For document generation, we use the prompts in Table~\ref{tab:rephrasing-prompt}, Table~\ref{tab:entity-extraction-prompt}, Table~\ref{tab:entity-linking-prompt}, Table~\ref{tab:strategy-prompt}, and Table~\ref{tab:active-reading-prompt}. For QA generation, we use a temperature of 1.0, a top-p of 1.0, and a maximum length of 2048. For document generation, we use a temperature of 0.7, a top-p of 0.95, and a maximum length of 4096. We use Llama 3.1 8B Instruct for QA generation in all experiments, including the generation of questions for Focal Rewriting. For the experiments in Section~\ref{subsec:gen}, we use Llama 3.1 70B Instruct for response and document generation, except on FinanceBench, where we use Qwen3 30B A3B Instruct for both response and document generation.

\begin{table}[t]
\centering
\begin{tabular}{p{0.95\textwidth}}
\toprule
\texttt{Generate question-answer pairs from the following article.} \\[0.5em]
\texttt{Article:} \\
\texttt{\{article\}} \\[0.5em]
\texttt{ONLY `Question: ...` and `Answer: ...` tags are allowed. DO NOT include any other text.} \\
\bottomrule
\caption{Prompt template used for question-answer pair generation from an article.}
\label{tab:qa-prompt}
\end{tabular}
\end{table}

\begin{table}[t]
\centering
\begin{tabular}{p{0.95\textwidth}}
\toprule
\texttt{Rewrite the following document to help the user understand the document better.} \\[0.5em]
\texttt{<document>} \\
\texttt{\{document\}} \\
\texttt{</document>} \\
\bottomrule
\caption{Prompt template used for document rephrasing~\citep{maini2024rephrasing}.}
\label{tab:rephrasing-prompt}
\end{tabular}
\end{table}

\begin{table}[t]
\centering
\begin{tabular}{p{0.95\textwidth}}
\toprule
\texttt{As a knowledge analyzer, your task is to dissect and understand an article provided by the user.} \\
\texttt{You are required to perform the following steps:} \\[0.5em]
\texttt{1. Summarize the Article: Provide a concise summary of the entire article, capturing the main points and themes.} \\
\texttt{2. Extract Entities: Identify and list all significant "nouns" or entities mentioned within the article. These entities should include but not limited to:} \\
\texttt{* People: Any individuals mentioned in the article, using the names or references provided.} \\
\texttt{* Places: Both specific locations and abstract spaces relevant to the content.} \\
\texttt{* Object: Any concrete object that is referenced by the provided content.} \\
\texttt{* Concepts: Any significant abstract ideas or themes that are central to the article's discussion.} \\[0.5em]
\texttt{Try to exhaust as many entities as possible. Your response should be structured in a JSON format to organize the information effectively.} \\
\texttt{Ensure that the summary is brief yet comprehensive, and the list of entities is detailed and accurate.} \\[0.5em]
\texttt{Here is the format you should use for your response:} \\
\texttt{\{\{} \\
\texttt{\ \ \ \ "summary": "<A concise summary of the article>",} \\
\texttt{\ \ \ \ "entities": ["entity1", "entity2", ...]} \\
\texttt{\}\}} \\[0.5em]
\texttt{Article:} \\
\texttt{\{document\}} \\
\bottomrule
\caption{Prompt template used to extract entities for EntiGraph~\citep{yangsynthetic}.}
\label{tab:entity-extraction-prompt}
\end{tabular}
\end{table}

\begin{table}[t]
\centering
\begin{tabular}{p{0.95\textwidth}}
\toprule
\texttt{You will act as a knowledge analyzer tasked with dissecting an article provided by the user. Your role involves two main objectives:} \\
\texttt{1. Rephrasing Content: The user will identify two specific entities mentioned in the article. You are required to rephrase the content of the article twice:} \\
\texttt{\ \ \ \ * Once, emphasizing the first entity.} \\
\texttt{\ \ \ \ * Again, emphasizing the second entity.} \\
\texttt{2. Analyzing Interactions: Discuss how the two specified entities interact within the context of the article.} \\[0.5em]
\texttt{Your responses should provide clear segregation between the rephrased content and the interaction analysis. Ensure each section of the output include sufficient context, ideally referencing the article's title to maintain clarity about the discussion's focus.} \\[0.5em]
\texttt{Here is the format you should follow for your response:} \\[0.5em]
\texttt{\#\#\# Discussion of <title> in relation to <entity1>} \\
\texttt{<Rephrased content focusing on the first entity>} \\[0.5em]
\texttt{\#\#\# Discussion of <title> in relation to <entity2>} \\
\texttt{<Rephrased content focusing on the second entity>} \\[0.5em]
\texttt{\#\#\# Discussion of Interaction between <entity1> and <entity2> in context of <title>} \\
\texttt{<Discussion on how the two entities interact within the article>} \\[0.5em]
\texttt{\#\#\# Document} \\
\texttt{\{document\}} \\[0.5em]
\texttt{\#\#\# Entities:} \\
\texttt{- \{entity1\}} \\
\texttt{- \{entity2\}} \\
\bottomrule
\caption{Prompt template used for entity linking for generating EntiGraph documents~\citep{yangsynthetic}.}
\label{tab:entity-linking-prompt}
\end{tabular}
\end{table}

\begin{table}[t]
\centering
\begin{tabular}{p{0.95\textwidth}}
\toprule
\texttt{Consider the following document. What are some strategies specific to this document that I can use to help me learn and remember all of the information contained? Use markdown and prefix each strategy with \#\#.} \\[0.5em]
\texttt{<document>} \\
\texttt{\{document\}} \\
\texttt{</document>} \\
\bottomrule
\caption{Prompt template used for generating active reading strategies~\citep{lin2025learning}.}
\label{tab:strategy-prompt}
\end{tabular}
\end{table}

\begin{table}[t]
\centering
\begin{tabular}{p{0.95\textwidth}}
\toprule
\texttt{Here's a learning strategy.} \\
\texttt{\{strategy\}} \\[0.5em]
\texttt{Apply this strategy to the following document:} \\[0.5em]
\texttt{<document>} \\
\texttt{\{document\}} \\
\texttt{</document>} \\
\bottomrule
\caption{Prompt template used for active reading document generation with a provided learning strategy~\citep{lin2025learning}.}
\label{tab:active-reading-prompt}
\end{tabular}
\end{table}

\begin{table}[t]
\centering
\begin{tabular}{p{0.95\textwidth}}
\toprule
\texttt{<document>} \\
\texttt{\{document\}} \\
\texttt{</document>} \\[0.5em]
\texttt{Here's a learning strategy.} \\
\texttt{\{strategy\}} \\[0.5em]
\texttt{Apply this strategy to the document above, with the focus on the question: \{query\}} \\
\bottomrule
\caption{Prompt template used for Focal Rewriting active reading with a provided learning strategy.}
\label{tab:query-guided-active-reading-prompt}
\end{tabular}
\end{table}

\section{Measuring Similarity of the Synthetic Data in Gradient Space}\label{sec:appendix-grad}

\begin{figure}[h!]
\centering
\includegraphics[width=0.6\linewidth]{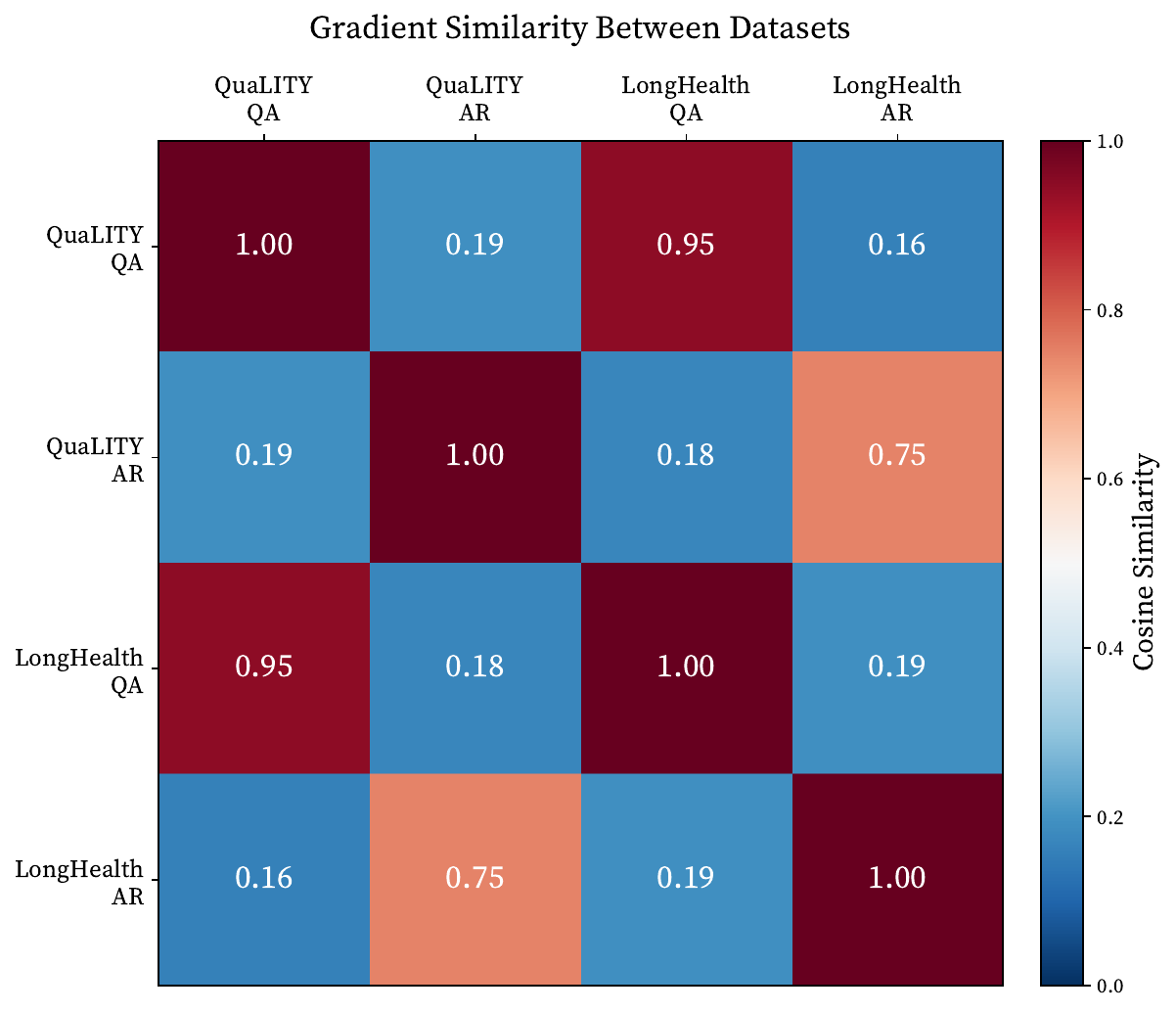}
\caption{\textbf{Average gradient similarity between datasets.} We compute gradient embeddings for each data point and use cosine similarity between embeddings to measure the similarity of gradients across datasets. (1) QA examples from different datasets exhibit high gradient similarity ($\geq 0.94$), whereas QA examples and documents (AR) from different datasets show low gradient similarity ($\leq 0.25$). (2) Even QA and documents from the same dataset do not exhibit high gradient similarity ($\leq 0.26$).}
\label{fig:grad_sim}
\end{figure}

To analyze how synthetic QA and synthetic documents are similar to each other, we compute gradient embeddings for synthetic QAs and synthetic documents (AR) generated from two datasets (QuaLITY, LongHealth~\citep{adams2025longhealth}), and measure both intra- and inter-set gradient-embedding similarity. All synthetic datasets are generated using 70B generator. 

Specifically, we follow \citet{jungprismatic} when computing the gradient embeddings: we use next-token prediction loss, Qwen3 0.6B as the model, and a Johnson-Lindenstrauss transform~\citep{johnson1984extensions} to reduce the gradient dimensionality. We sample 16 batches with the sequence length of 2048 from each data types.

Figure~\ref{fig:grad_sim} shows the results. QA datasets exhibit high gradient similarity across different domains, whereas AR datasets show lower gradient similarity. However, QA and AR data from the same domain exhibit the lowest gradient similarity, suggesting that data type is an important factor in shaping training signals.

\section{More results}\label{sec:appendix-more}

\subsection{Measuring diversity of generated documents}

We measure diversity from two perspectives: semantic diversity and lexical diversity. For semantic diversity, we use the Vendi Score~\citep{friedmanvendi}, which quantifies how varied the data instances are. Specifically, we compute an embedding for each instance using Qwen3-Embedding-8B and then use the cosine similarity between embeddings to construct the pairwise similarity matrix required for the Vendi Score. For lexical diversity, we report the unique 4-gram ratio, which captures the extent of surface-form variation in the text.

\begin{figure}[t!]
  \centering
  \begin{subfigure}[t]{0.49\columnwidth}
    \centering
    \includegraphics[width=\linewidth]{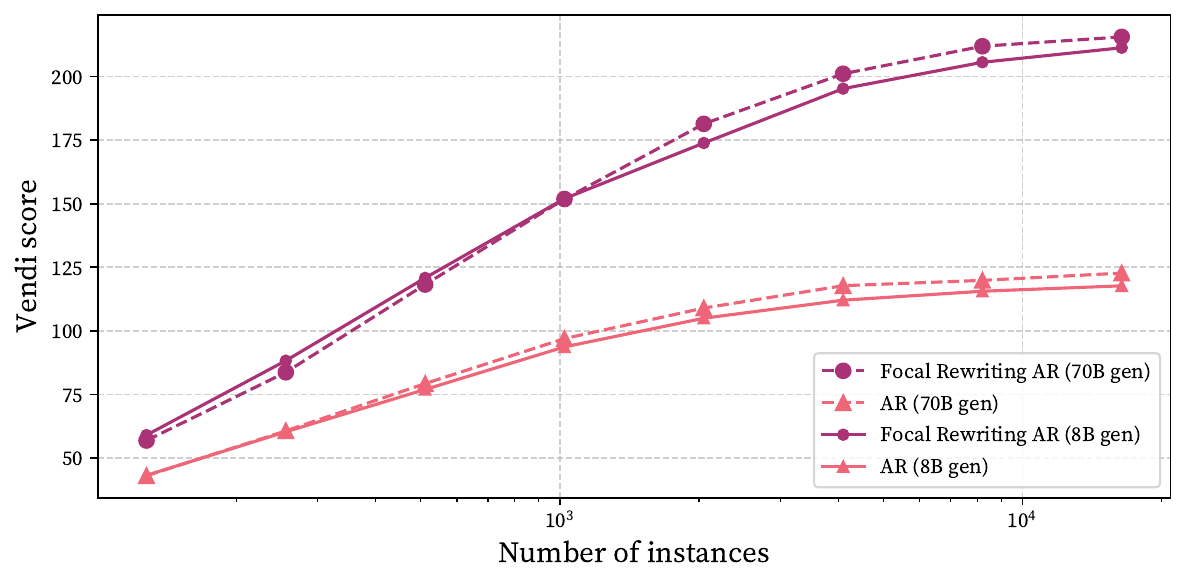}
    \label{fig:emb-vendi}
  \end{subfigure}\hfill
  \begin{subfigure}[t]{0.49\columnwidth}
    \centering
    \includegraphics[width=\linewidth]{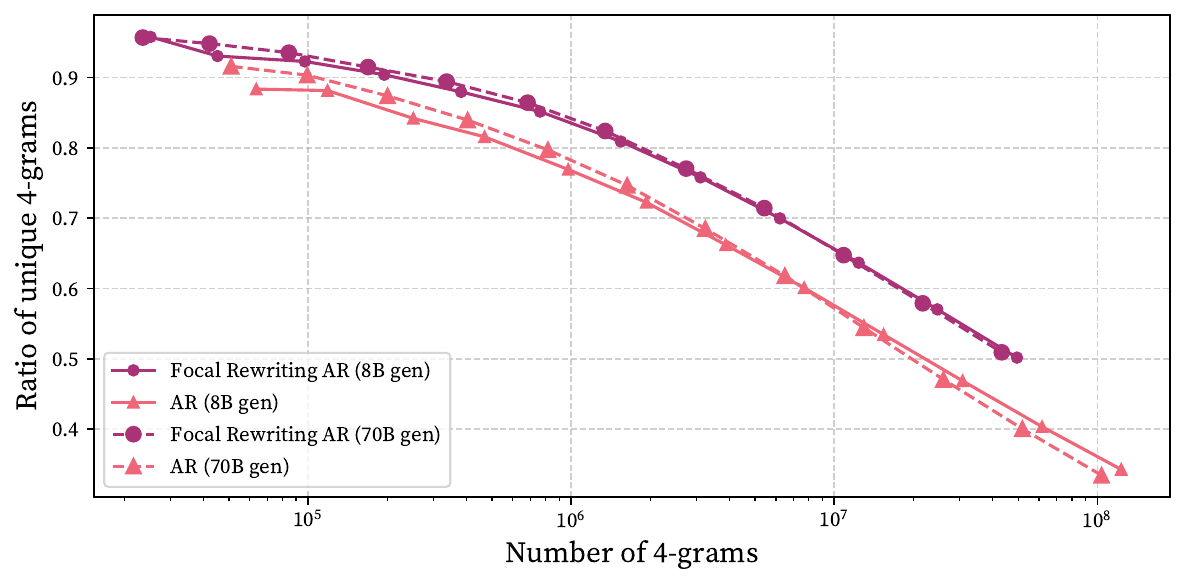}
    \label{fig:lexical-div}
  \end{subfigure}
    \caption{\textbf{How data diversity changes with different synthetic document generation methods.} (Left) Semantic diversity of synthetic documents, measured by the Vendi score~\citep{friedmanvendi} using embedding-based similarity to compute distances between data points. (Right) Lexical diversity of synthetic documents, measured as the ratio of unique 4-grams in the data. For both metrics, higher values indicate greater diversity. (1) Focal Rewriting increases both semantic and lexical diversity at all dataset sizes, and (2) scaling the generator does not significantly affect diversity.}
  \label{fig:diversity}
\end{figure}

In Figure~\ref{fig:diversity}, we show how the diversity of AR documents and Focal Rewriting AR documents changes across different data sizes on QuaLITY. We plot the results this way to compare how diversity changes under different data budgets. The figure shows that Focal Rewriting yields higher lexical and semantic diversity. Interestingly, using a stronger model does not lead to higher diversity.

\subsection{Finding optimal mixing ratio for synth mixed training}
\begin{figure}[t!]
\centering
\includegraphics[width=0.7\linewidth]{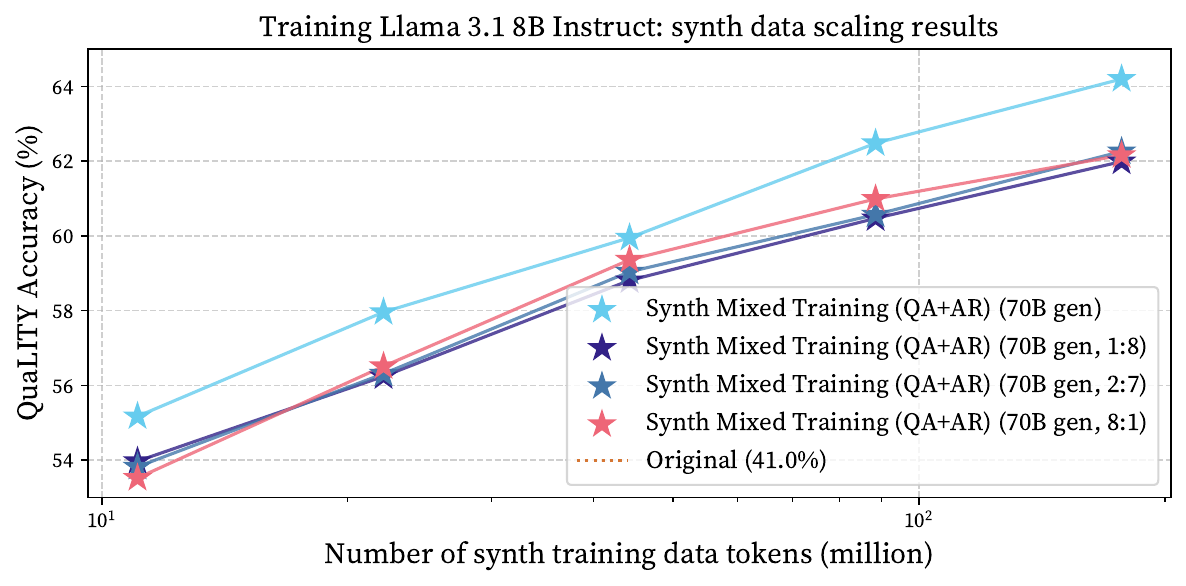}
\caption{\textbf{Synthetic mixed training with different synthetic QA-document mixing ratios.} We test four mixing ratios of QA and AR: (1:1), (1:8), (2:7), and (8:1). The remaining 10\% is used for replay with FineWeb. Using 1:1 mixing gives the best result.}
\label{fig:llama-mix-ratio}
\end{figure}

We experiment with different mixing ratios of synthetic QA data and documents for mixed training. Figure~\ref{fig:llama-mix-ratio} shows the results of training Llama 3.1 8B on data generated by the 70B model. Among the tested variants, a 1:1 mixing ratio yields the best performance.

\section{Additional Related Works}
\paragraph{Parameter-efficient training for new knowledge.} Parameter-efficient adaptation is a promising direction for teaching models new knowledge. LoRA~\citep{hulora} has been widely used to adapt models through low-rank updates to their weights, and, combined with context distillation~\citep{snell2022learning}, \citet{cacciatraining} propose training LoRA layers to acquire new knowledge. However, their Llama 8B model trained on QuaLITY achieves 59.3\% accuracy, which remains substantially below our results. \citet{bidermanlora} also show that low-rank updates can limit the acquisition of new knowledge, highlighting a key limitation of LoRA for knowledge-intensive learning.

Motivated by the hypothesis that Transformer key-value (KV) caches function as a form of knowledge base~\citep{geva2021transformer}, \citet{eyuboglu2025cartridges} propose an end-to-end training approach that optimizes only the KV cache to store knowledge. More recently, \citet{zweiger2026fast} introduce an optimization method that updates the KV cache to compress knowledge without requiring end-to-end training. Although we view these approaches as promising, we do not include them as baselines for two reasons. First, applying these compression-based methods to our setting is infeasible because concatenating all documents would require context lengths of 1.6M and 4.8M tokens, respectively, which are not supported by the base model we use. Second, even if the base model supported context lengths beyond 1M tokens, their performance degrades substantially at high compression ratios (i.e., when compressing by more than $20\times$), making them difficult to apply in our setting. Consistent with these limitations, \citet{zweiger2026fast} evaluate on QuaLITY by compressing only a single document, while on LongHealth, \citet{zweiger2026fast} and \citet{eyuboglu2025cartridges} compress only five and ten documents, respectively. In contrast, we train models on all documents in each dataset: 265 documents from QuaLITY and 400 documents from LongHealth.

\paragraph{Alleviating forgetting.} Continued training often leads to the forgetting of existing knowledge in language models. A common technique for mitigating this issue is replay--reusing pretraining data during the continued training stage~\citep{lin2025learning, yangsynthetic, kotha2026replaying, liu2025midtraining}. In a symbolic distillation setup, \citet{agarwal2024policy, lu2025onpolicydistillation} suggest using on-policy distillation: compared to training the model with the synthetic data generated by another model (e.g., stronger model), using the self-generated data for the points to compute the loss leads to less forgetting while learning new knowledge well.

There have also been attempts to address forgetting through improved language model architectures. For example, \citet{lin2025continual} suggest using memory layers~\citep{he2024mixture, berges2024memory}, which are identical to Mixture-of-Experts~\citep{shazeer2017outrageously} models but use a large number of experts in a specific layer. These layers are updated specifically for new knowledge, reducing interference with existing knowledge. The paper shows that there is a trade-off between learning new knowledge and forgetting existing knowledge, and that sparse model updates with memory layers provide a better Pareto frontier than full fine-tuning or LoRA~\citep{bidermanlora, hulora}.

\end{document}